\documentclass[journal]{IEEEtran}

\usepackage{amsmath,amsfonts}
\usepackage{algorithmic}
\usepackage{algorithm}
\usepackage{stfloats}
\usepackage{url}
\usepackage{graphicx}
\usepackage{cite}
\usepackage{subfigure}
\usepackage{xcolor}
\usepackage{bm}

\begin{document}

\title{DGSense: A Domain Generalization Framework for Wireless Sensing}

\author{Rui Zhou, Yu Cheng, Songlin Li, Hongwang Zhang, Chenxu Liu
\thanks{The authors are within the University of Electronic Science and Technology of China.}}

\maketitle

\begin{abstract}
Wireless sensing is of great benefits to our daily lives. By analyzing the impact of human body and movements on wireless propagation, a variety of sensing tasks can be enabled. However, wireless signals are sensitive to the surroundings. Various factors, e.g. environments, locations, and individuals, may induce extra impact on wireless propagation. Such a change can be regarded as a domain, in which the data distribution shifts. A vast majority of the sensing schemes are learning-based. They are dependent on the training domains, resulting in performance degradation in unseen domains. Researchers have proposed various solutions to address this issue. But these solutions leverage either semi-supervised or unsupervised domain adaptation techniques. They still require some data in the target domains and do not perform well in unseen domains. In this paper, we propose a domain generalization framework DGSense, to eliminate the domain dependence problem in wireless sensing. The framework is a general solution working across diverse sensing tasks and wireless technologies. Once the sensing model is built, it can generalize to unseen domains without any data from the target domain. To achieve the goal, we first increase the diversity of the training set by a virtual data generator, and then extract the domain independent features via episodic training between the main feature extractor and the domain feature extractors. The feature extractors employ a pre-trained Residual Network (ResNet) with an attention mechanism for spatial features, and a 1D Convolutional Neural Network (1DCNN) for temporal features. To demonstrate the effectiveness and generality of DGSense, we evaluated on WiFi gesture recognition, Millimeter Wave (mmWave) activity recognition, and acoustic fall detection. All the systems exhibited high generalization capability to unseen domains, including new users, locations, and environments, free of new data and retraining. 
\end{abstract}


\section{Introduction}
Human sensing plays an important role in many fields of daily life. Conventional approaches to human sensing are based on visions or wearables. Vision-based solutions deploy cameras to collect and analyze visual data. They can achieve high accuracy, but require Line of Sight (LOS) and luminous lighting to work properly. In addition, cameras impose privacy concerns, hence they are not appropriate to be deployed in private spaces. Human sensing based on wearables eliminates the limitations of LOS and lighting and does not impose privacy concerns. These solutions exploit the sensors embedded in the wearables to realize sensing. However, wearable-based solutions require the users to wear dedicated devices continuously, causing inconvenience and reluctance.  
To overcome the shortcomings of cameras and wearables, human sensing has shed light on wireless signals. The ubiquitous availability of wireless signals brings new opportunities to human sensing. The past two decades have witnessed extensive wireless sensing techniques, including but not limited to Radio Frequency (RF)~\cite{Sigg:2013,Orphomma:2013}, WiFi~\cite{MaYS:2019,LiuJ:2020}, Millimeter Wave (mmWave)~\cite{LiuH:2020,ZhaoP:2021}, Long Range (LoRa)~\cite{Bashima:2019,ZhangF:2020}, and acoustic~\cite{CaiC:2022,WangY:2022}. In an environment covered with wireless signals, the presence of human bodies alters the propagation of wireless signals, causing them to carry rich user-specific information, which can be exploited to infer human contexts. Wireless sensing avoids the inconvenience brought by wearables and eliminates the privacy invasion from cameras. A plethora of studies have been carried out on wireless sensing over the years. 

However, wireless signals are sensitive to the surroundings. Factors such as environments, locations, and individuals may induce extra impact on wireless propagation and cause the data distributions to shift. Each such a change can be regarded as a domain. Although the sensing model can achieve impressive performance in the training domains, they often suffer from performance degradation in unseen domains. To deploy wireless sensing in real-world applications, the domain dependence problem must be solved. The domain where the model is trained is called the source domain, whereas the domain where the model is applied/tested is called the target domain. If all the data in the target domain are not previously encountered, we call it the unseen domain or new domain. Existing solutions to the domain dependence issue typically leverage semi-supervised or unsupervised domain adaptation methods, requiring partially labeled or unlabeled data from the target domain to transfer the sensing model from the source to the target domain. However, collecting data from the target domain beforehand is often infeasible in practice. We need a solution that can automatically generalize the sensing model to new/unseen domains without any new data and retraining. This is a domain generalization (DG) approach.

In this paper, we propose a domain generalization framework for wireless sensing---DGSense. It aims to enable generalization to new/unseen domains without any data from the new domains, meanwhile the framework is general to diverse sensing tasks on diverse wireless technologies. The key points in the framework are: (1) \textbf{a cross-modal virtual data generator}, to increase the diversity of the training set and enhance the robustness of the sensing model; (2) \textbf{a spatial-temporal feature extractor}, based on a pre-trained Residual Network (ResNet) with an attention mechanism for spatial features and a 1D Convolutional Neural Network (1DCNN) for temporal features; (3) \textbf{an episodic training strategy}, to extract the domain independent features for the purpose of domain generalization. We evaluated DGSense on WiFi gesture recognition, mmWave activity recognition, and acoustic fall detection. The results demonstrated the effectiveness of DGSense.
The contributions of the paper are as follows.
\begin{itemize} 
\item Solves the domain dependence problem in wireless sensing, by proposing a general domain generalization framework DGSense. Based on DGSense, we implemented a WiFi gesture recognition system, an mmWave activity recognition system, and an acoustic fall detection system, all achieving domain independence and generalization to unseen domains.
\item Proposes a cross-modal virtual data generator based on Variational Autoencoder (VAE), which synthesizes cross-modal virtual data and meanwhile ensures the consistency between the multiple modalities in the same data. It alleviates data scarcity, enhances data diversity, and helps improve model robustness. 
\item Applies an episodic training strategy to extract domain independent features. The strategy is composed of a main network and several domain networks associating with each domain. Each network consists of a feature extractor and a classifier. Through episodic training between the main network and each domain network, the main feature extractor acquires the capability of extracting domain independent features, which are classified by the main classifier to achieve robust sensing. 
\item Designs a spatial-temporal feature extractor, which employs a 1DCNN to extract the temporal features and a ResNet with attention to extract the spatial features. The feature extractor is applicable to diverse wireless signals, proved by WiFi, mmWave and acoustic data.
\end{itemize}

\section{Related work}
\label{SecRelatedWork}

\subsection{Robust wireless sensing}
\subsubsection{WiFi sensing}
WiFi sensing has been extensively studied since the release of Channel State Information (CSI) tools~\cite{Halperin:2010,XieY:2015}. Many solutions have been proposed to improve the robustness. 
Exploiting adversarial networks, EI~\cite{JiangWJ:2018} could remove environment and user-specific information and learn transferable features of activities. 
CsiGAN~\cite{XiaoCJ:2019} enabled activity recognition adaptive to users based on semi-supervised Generative Adversarial Network (GAN), leveraging a complement generator to produce diverse fake samples to train a robust discriminator.
Kang et al.~\cite{KangH:2021} realized cross domain gesture recognition based on CSI Doppler Frequency Shift (DFS) through multi-source unsupervised domain adaptation, by applying adversarial learning and feature disentanglement to remove gesture irrelevant factors. 
OneFi~\cite{XiaoR:2021} could recognize unseen gestures with only one (or few) labeled samples. It utilized virtual gesture generation to reduce the effort in data collection and employed a one-shot learning framework based on transductive fine-tuning to eliminate model retraining. 
These methods required some labeled or unlabeled data from the target domains for model training, while our method did not require any target domain data for model training.
Widar3.0~\cite{ZhangY:2022} achieved cross domain gesture recognition by extracting the domain independent feature Body-coordinate Velocity Profile (BVP) from CSI DFS. Widar3.0 required one transmitter and at least three receivers to work, while our method required only one transmitter and one receiver.  

\subsubsection{mmWave sensing}
Due to high frequency and wide bandwidth, mmWave can be used to realize highly accurate sensing. Researchers have developed various mmWave-based sensing applications, mostly exploiting Frequency Modulated Continuous Waves (FMCW). 
S. Wang et al.~\cite{Soli:2016} built a gesture recognition system based on Google’s Soli sensor, leveraging a combination of deep convolutional and recurrent neural networks.
Zhang et al.~\cite{ZhangG:2020} processed range-Doppler maps (RDM) with a zero-filling strategy to boost the range and velocity information of gestures and constructed a 3DCNN model and a CNN-Long Short Term Memory (LSTM) model to reveal the temporal gesture signatures encoded in multiple frames. 
Amin et al.~\cite{Amin2020} classified daily activities, in particular contiguous motions, using micro-Doppler signatures and range maps. 
Chen et al.~\cite{ChenH:2022} designed a temporal 3DCNN to deal with a series of range-Doppler maps and added a temporal attention module to emphasize the sequential relation between each frame. 
These methods focused on in-domain sensing without considering the cross domain issue.
mHomeGes~\cite{LiuH:2020} proposed concentrated position-Doppler profile (CPDP) to represent the unique features from different arm joints and recognized continuous arm gestures based on a lightweight CNN. mHomeGes could work across various smart-home scenarios regardless of the impact of surrounding interference, but domain independence was not under consideration of this work.  
M-Gesture~\cite{LiuH:2022} achieved person-independent gesture recognition by incorporating a pseudo representative model (PRM) and a custom-built neural network to depict and extract the inherent gesture features. Their experiments showed that M-Gesture required about 20 persons to achieve high accuracy for new users and the changes in user orientations still degraded the performance. 
 
\subsubsection{Acoustic sensing}
Acoustic signals can be used for human sensing. A variety of applications have been developed in recent years~\cite{CaiC:2022}. 
DeepRange~\cite{MaoW:2020} synthesized training data and realized acoustic ranging based on deep learning. 
EchoSpot~\cite{LianJEchoSpot:2021} localized a person by periodically emitting acoustic chirps, capturing the reflections from the human body and the walls, and normalizing their cross-correlation. 
EarIO~\cite{LiK:2022} extracted the depth features of the face and tracked the facial expressions via cross-correlation between emitted and reflected acoustic signals. To adapt to new users, EarIO required a small amount of data from new users.
Lian et al.~\cite{LianJ:2021} detected falls by emitting continuous waves of fixed frequency and calculating Doppler frequency spectrogram of the reflections. Singular Value Decomposition (SVD) and Hidden Markov Model (HMM) were applied to classify the motions as falls or non-falls. 
StruGesture~\cite{WangL:2021} exploited the structure-borne sounds to classify gestures on the back surface of mobile phones. It leveraged deep adversarial learning to learn the gesture-specific representation, and required a few samples from the new users to achieve proper recognition for new users.

\subsection{Domain adaptation and generalization}
\subsubsection{Domain adaptation}
Domain adaptation (DA) aims to utilize the data in the source domains to solve the learning problem in the target domains. Some domain adaptation methods have been proposed to tackle the domain dependence problem in wireless sensing.
Feature alignment is a common method of domain adaptation. The features in the source and the target domains are mapped to a shared space, in which the distribution divergence between them is minimized, so that the sensing model built in the source domain can then be applied to the target domain. AdapLoc~\cite{ZhouR:2021} achieved adaptive CSI localization by mapping the source and the target domains into a shared space and minimizing the distance between the fingerprints at the same location and maximizing the distance between the fingerprints at different locations. DANGR~\cite{HanZ:2020} recognized gestures using CSI based on a deep adaptation network. To shrink the domain discrepancy, DANGR adopted Multi-Kernel Maximal Mean Discrepancy (MK-MMD).
Researchers also proposed synthesizing virtual samples to enlarge the training set and increase the diversity, thereby enhancing the robustness. FiDo~\cite{ChenX:2020} could localize diverse users with labeled CSI data from one or two users, leveraging a data augmenter that introduced data diversity using VAE and a domain adaptive classifier. Zhang et al.~\cite{ZhangJ:2021} synthesized variant activity data through CSI transformation to mitigate the impact of activity inconsistency and subject-specific issue for CSI activity recognition. 
Domain adaptation can also be achieved by adversarial networks. Regarding the target domain data as fake samples, the domain independent features can be extracted through adversarial training. By exploiting adversarial networks, EI~\cite{JiangWJ:2018} could remove environment and user-specific information and achieve environment independent activity recognition based on CSI. CsiGAN~\cite{XiaoCJ:2019} enabled CSI activity recognition adaptive to users based on semi-supervised GAN. 
Domain adaptation methods require extra efforts to collect data from the target domains. They have to retrain the classifiers each time new target domains are added. Therefore, they have difficulties in generalizing to new/unseen domains.  

\subsubsection{Domain generalization}
Domain generalization extends domain adaptation by generalizing to the target domains without any target domain data and retraining. It focuses on learning a domain independent sensing model from the source domains and achieving high performance in unseen target domains~\cite{WangJ:2022}. Domain generalization is still at its early stage and most researches are on image recognition. H. Li et al.~\cite{LiH:2018} applied adversarial Autoencoders and MMD to align feature distributions and extract domain independent features. D. Li et al.~\cite{LiD:2019} enhanced model robustness to new domains through episodic training, exposing the model to diverse samples. Qiao et al.~\cite{QiaoF:2020} combined meta-learning and Wasserstein Autoencoders to generate virtual data, improving generalization capabilities.   

In the field of wireless sensing, general domain generalization methods for diverse sensing tasks on top of diverse wireless signals are still lacking. Previous studies either exploit domain adaptation methods requiring some target domain data, or design specific signal processing techniques for specific sensing tasks. The main contribution of our work is a general domain generalization framework for wireless sensing tasks.

\section{Framework of domain generalization}
\label{SecFramework}

\subsection{Problem formulation}
Suppose $\mathcal D^s$ represents the set of source domains. A domain refers to a different scenario, e.g. an individual, a location, an environment, etc. $\mathcal D^s$ can be expressed as: 
\begin{equation}
\mathcal D^s = \{ D_{1}^{s}, D_{2}^{s}, \cdots, D_{N}^{s} \}
\end{equation}
where $N$ is the number of source domains. Each source domain contains multiple samples and their labels, which can be expressed as: 
\begin{equation}
D_{i}^{s} = \{(x_{ij}^{s}, y_{ij}^{s})|j=1,2,\cdots,n_i\}
\end{equation}
where $x_{ij}^{s}$ represents sample $j$ in domain $i$, $y_{ij}^{s}$ represents its label, $n_i$ is the number of samples in domain $i$. Suppose $D^s$ denotes the training set containing all the training samples, expressed as:
\begin{equation}
D^{s} = \{(x_{l}^{s}, y_{l}^{s})|l=1,2,\cdots,n\}
\end{equation}
where $n=\sum_{i=1}^N{n_i}$ is the number of all the training samples, $x_{l}^{s}$ represents a sample, and $y_{l}^{s}$ represents its label. Suppose $D^{t}$ represents the testing set, i.e. the unseen target domain. The domain generalization problem can be described as: we train the sensing model with the source domain data $D^{s}$ and generalize it to the unseen target domain $D^{t}$, without any target domain data participating in the training. 

\subsection{Framework overview}
\begin{figure*}
\centerline{\includegraphics[width=16cm]{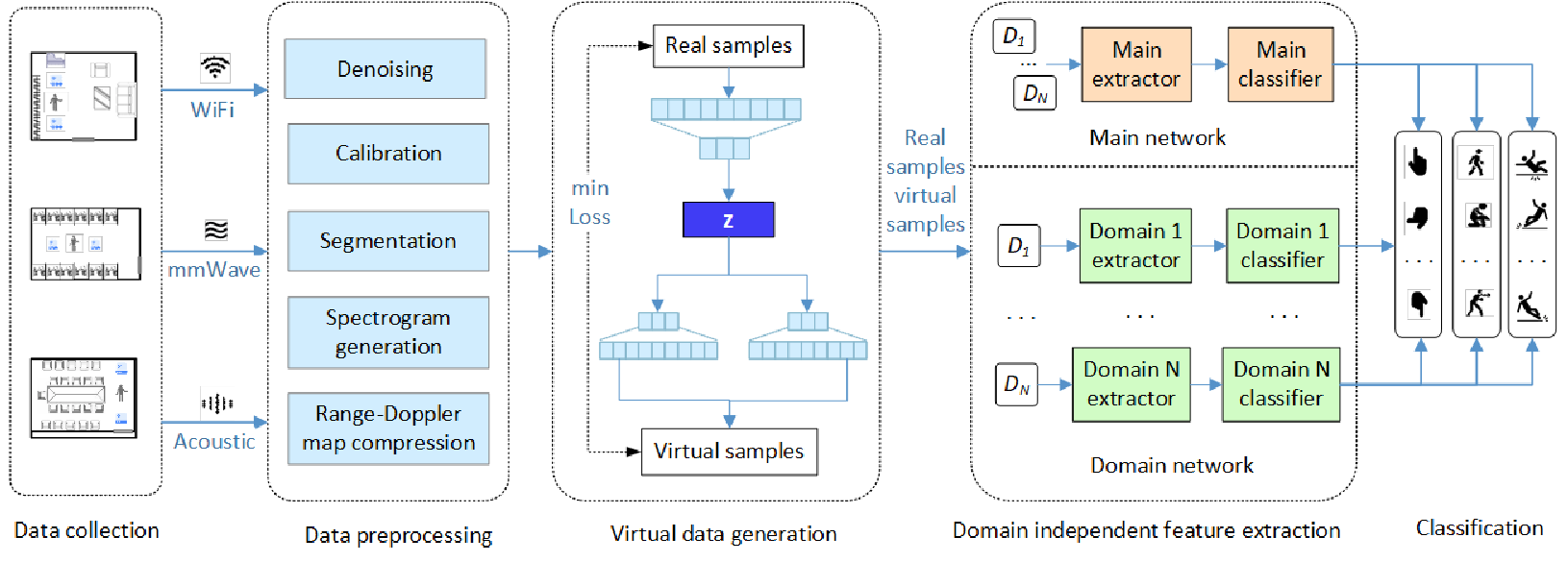}}
\caption{The domain generalization framework (DGSense).}
\label{FigFramework}
\end{figure*}

We propose a framework of domain generalization for wireless sensing---DGSense, to mitigate the impact of domain diversity and achieve domain independent sensing. DGSense enables the sensing model trained in the source domains to sustain high performance in the unseen target domains. It is a general framework, accommodating diverse sensing tasks on top of diverse wireless signals. Fig.~\ref{FigFramework} depicts the overall framework, composed of four main components: \textit{data collection}, \textit{data preprocessing}, \textit{virtual data generation}, \textit{domain independent feature extraction and classification}. The virtual data generation component augments the training set with more diversity. It generates the virtual data to augment the source domains, allowing the sensing model trained in the source domains to be more robust. The domain independent feature extraction and classification component concentrates on extracting the sensing-related features whilst mitigating the influence of environmental factors. It consists of a main network and multiple domain networks. Each domain network is associated with a source domain. Through episodic training between the main network and the domain networks, the main network acquires the capability of extracting domain independent features. The domain networks are employed only in the training phase, while in the testing phase only the main network is involved.

\subsection{Data collection and preprocessing}
The framework supports various wireless technologies, such as WiFi, mmWave and acoustic. The data collection methods rely on the wireless technologies and the preprocessing methods adapt to the data. The raw WiFi data are time-series of amplitude and phase, which need to be denoised, calibrated and segmented. The raw mmWave data are multi-frame range-Doppler maps, which are compressed to reduce the complexity. The raw acoustic data are sound waves, which are denoised and transformed to Doppler spectrograms. To reduce the noise in time-series data, the moving average algorithm or the median filter can be applied to smooth the data and eliminate the high-frequency noise. To reduce the noise in images, the threshold filtering algorithm can be applied. After that, the data are segmented to obtain the relevant sensing part. For time-series data, the variance threshold algorithm can be leveraged to identify the part with the most significant fluctuations. For images, the relevant segments can be located by the Power Burst Curve (PBC) method. 

\subsection{Virtual data generation}
To increase data diversity and reduce the effort of data collection, we incorporate a virtual data generator to expand the training set. The virtual data generator is based on VAE. In the training phase, we use the source domain data to train the virtual data generator. In the generating phase, the source domain data are input to the encoder to obtain the intermediate features, to which the identically distributed noise is added. The intermediate features with noise are passed to the decoder to obtain the reconstructions as the virtual data. 

\subsubsection{Single-modal generation}
\begin{figure*}
    \centering
    \subfigure[Single-modal]{
		\label{FigVirtualDataGenerator:a}
    \raisebox{0.5cm}{\includegraphics[width=6.5cm]{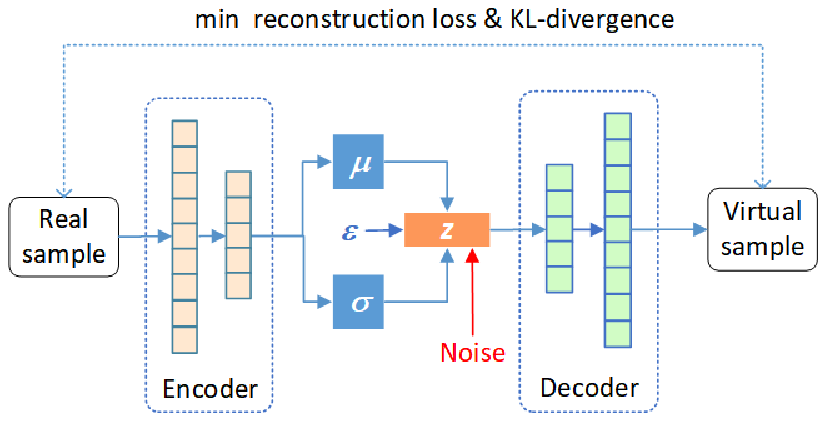}}}
		\hspace{1cm}
    \subfigure[Cross-modal]{
		\label{FigVirtualDataGenerator:b}
    \includegraphics[width=8cm]{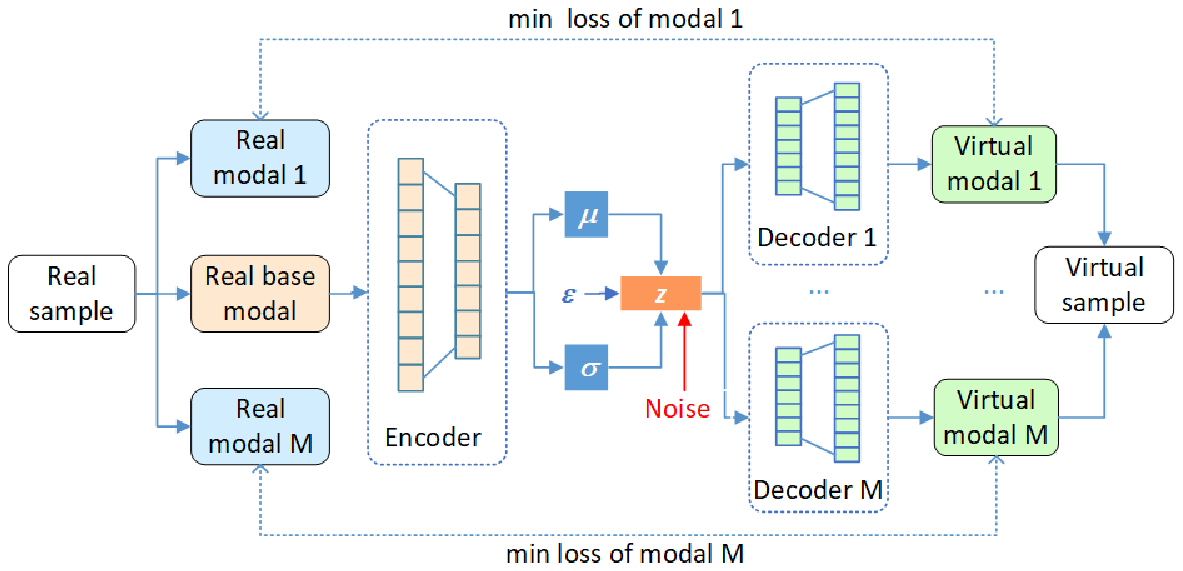}}
\caption{The virtual data generator.}
\label{FigVirtualDataGenerator}
\end{figure*}

For samples with a single modality, such as Doppler spectrograms, we construct a single-modal virtual data generator. Its structure is visualized in Fig.~\ref{FigVirtualDataGenerator:a}. The encoder is a CNN and the decoder is a deconvolutional neural network (DCNN). To train the generator, we input the real source domain data into the encoder and derive the reconstructions. To optimize the model parameters, we minimize both the reconstruction loss between the input data and the reconstructions and the Kullback–Leibler divergence (KL-divergence) between the intermediate feature distribution and the normal distribution. Assuming the input data is $x_{i}^{s}$, the intermediate feature $z_i$ is calculated by the encoder as:
\begin{equation}
    z_i = Encoder(x_i^s) = \mu_i + \sigma_i \odot \varepsilon_i 
\end{equation}
where $\mu_i$ is the mean value of the encoder output, $\sigma_i$ is the standard deviation of the encoder output, $\varepsilon_i \sim N(0,I)$ is a normal distribution, and $I$ represents identity matrix. The intermediate feature $z_i$ is passed to the decoder to obtain the reconstruction $x _{i}^{rec}$. The reconstruction loss between $x_{i}^{s}$ and $x _{i}^{rec}$ is minimized, along with a regularization term restricting the intermediate feature $z_i$ to a normal distribution. The loss function of the virtual data generator is defined as:
\begin{equation}
\footnotesize
	{\rm min}\ L_{VAE} = \frac{1}{n}\sum_{i=1}^{n} \left( MSE(x_{i}^{s}, x_{i}^{rec}) + \lambda \cdot KL(N(\mu_i,\sigma_i)||N(0,I)) \right) 
\end{equation}

For generating the virtual data, the real data $x _{i}^{s}$ is fed to the encoder to extract the intermediate feature $z_i$, and the noise with the identical distribution is added to $z_i$ in a specific proportion. The noised feature ${\tilde{z_i}}$ is then fed to the decoder to obtain the virtual data $x_{i}^{v}$: 
\begin{equation}
\begin{split}
		z_i	&=  Encoder(x_{i}^{s})	\\
		\tilde{z_i} &= \omega_{1} \cdot z_i + \omega_{2} \cdot noise \\
    x_{i}^{v} &= Decoder(\tilde{z_i})
\end{split}
\end{equation}
The labeled virtual data is thus generated in the form of $(x_{i}^{v}, y_{i}^{v})$, where $y_{i}^{v} =y_{i}^{s}$.

\subsubsection{Cross-modal generation}
When dealing with the samples containing multiple correlated modalities, such as WiFi samples, where amplitude, phase and spectrogram are correlated and used concurrently, it is necessary to maintain the consistency among the multiple modalities. When generating a virtual sample, a real sample is encoded to obtain the intermediate feature. The noise with the identical distribution is added to the intermediate feature, which is then decoded to obtain the virtual sample. As different modalities have different data distributions, the noises added to them also have different distributions. This will bring external inconsistency to these reconstructed modalities. To alleviate this problem, we introduce a cross-modal virtual data generator, leveraging the inter-modal relationships, as depicted in Fig.~\ref{FigVirtualDataGenerator:b}. The encoder is a CNN and the decoders are DCNNs. We first select one modality as the base modality, which has relationship with every other modality and acts as the base of generating other modalities. For virtual data generation, we input solely the base modality of the real sample, and the cross-modal generator will generate all the virtual modalities from the base modality. Hence in the cross-modal generator, there is only one encoder, but multiple decoders. Each decoder is associated with a modality. As all the virtual modalities stem from the base modality, they have consistency to form the virtual sample. Assume a training sample is denoted as $x_i^s=((x_i^s(1),x_i^s(2),\cdots,x_i^s(M)), y_i^{s})$, where $x_i^s(k)$ denotes the $k$-th modality, and $M$ is the number of all modalities. Assume $x_i^s(b)$ is the base modality data. During the training of the cross-modal generator, the base modality data $x_i^s(b)$ is input to the encoder to derive the intermediate feature $z_i$, which is then passed to the decoder of each modality to generate the reconstruction of that modality. Through training, the generator establishes a mapping between each modality with the base modality. The loss function of each modality can be defined as:
\begin{equation}
\footnotesize
 {\rm min}\ L_k  = \frac{1}{n} \sum_{i=1}^{n} \left( MSE(x_i^s(k), x_i^{res}(k)) + \lambda \cdot KL(N(\mu_i,\sigma_i)||N(0,I)) \right)
\end{equation}
The overall loss function of the generator can be defined as:
\begin{equation}
  {\rm min}\quad L_{VAE} = \sum_{k=1}^{M}L_k	
\end{equation}

During the generation phase, the real base modality data $x_i^s(b)$ is input to the encoder to obtain the intermediate feature $z_i$, and the noise with the identical distribution is added to $z_i$. The noised feature is then passed to each decoder to obtain the virtual data of that modality, which constitutes the complete virtual data:
\begin{equation}
\begin{split}
    & z_i = Encoder(x_i^s(b))	\\
    & \tilde{z_i} = \omega_{1} \cdot z_i + \omega_{2} \cdot noise  \\
    & x_i^v(k) = Decoder_{k}(\tilde{z_i})	\\
		& x_i^v=(x_i^v(1),x_i^v(2),\cdots,x_i^v(M))
\end{split}
\end{equation}
The virtual data from all the modalities constitute the complete virtual sample as $(x_i^v, y_i^{v})$, where $y_{i}^{v} = y_{i}^{s}$.

To leverage the cross-modal generator, the multiple modalities should be correlated. For WiFi sensing, the amplitude and the phase are from the same complex values, the spectrogram is generated from the amplitude time series, hence they have inter-modal relationships. Based on the inter-modal relationships, we can build the cross-modal generator. If there are no inter-modal relationships, we may build a multi-modal generator, which encodes and decodes each modality respectively and fuses them as one sample.  

\subsection{Domain independent feature extraction}
To achieve robust sensing, the key is to extract domain independent features. We adopt the episodic training strategy introduced by Li et al.~\cite{LiD:2019} for computer vision. The strategy encompasses a main network and multiple domain networks. The training process first trains the domain networks and then trains the main network. Through episodic training, the main network acquires the generalization capability to unseen domain finally.

\subsubsection{Network structure}
The episodic training strategy consists of a main network and multiple domain networks, as depicted in Fig.~\ref{FigEpisodic:a}. The main network is composed of a main feature extractor and a main classifier. Each domain network is associated with a source domain, consisting of a domain feature extractor and a domain classifier. All the feature extractors share the same structure and all the classifiers share the same structure, tailored to the wireless data. For time-series, we employ 1DCNN as the feature extractor to capture the temporal features. For images, we employ ResNet18~\cite{HeK:2016} with Convolutional Block Attention Module (CBAM)~\cite{Woo:2018} as the feature extractor to capture the spatial features. All the classifiers are fully-connected neural networks. 

\subsubsection{Training of the domain network}
\begin{figure*}
    \centering
    \subfigure[Network structure]{
		\label{FigEpisodic:a}
    \includegraphics[height=4cm]{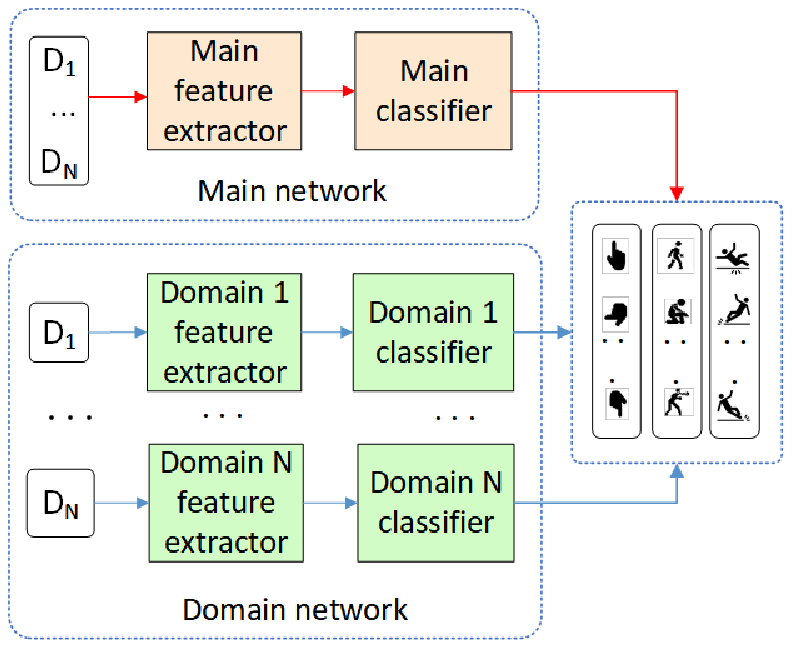}}
		\hspace{0.0cm}
    \subfigure[Training of domain networks]{
		\label{FigEpisodic:b}
    \includegraphics[height=4cm]{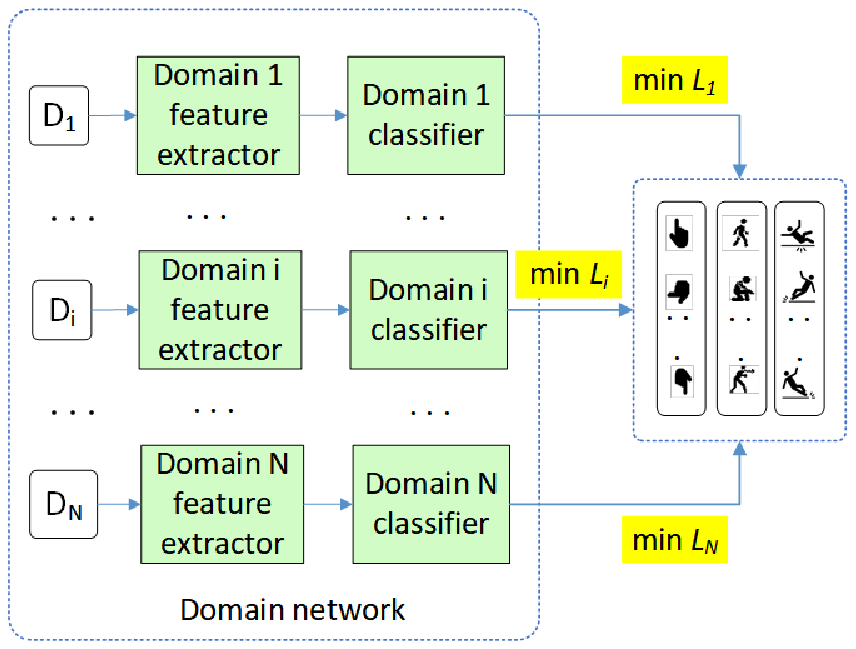}}
		\hspace{0.0cm}
    \subfigure[Training of main network]{
		\label{FigEpisodic:c}
    \includegraphics[height=4cm]{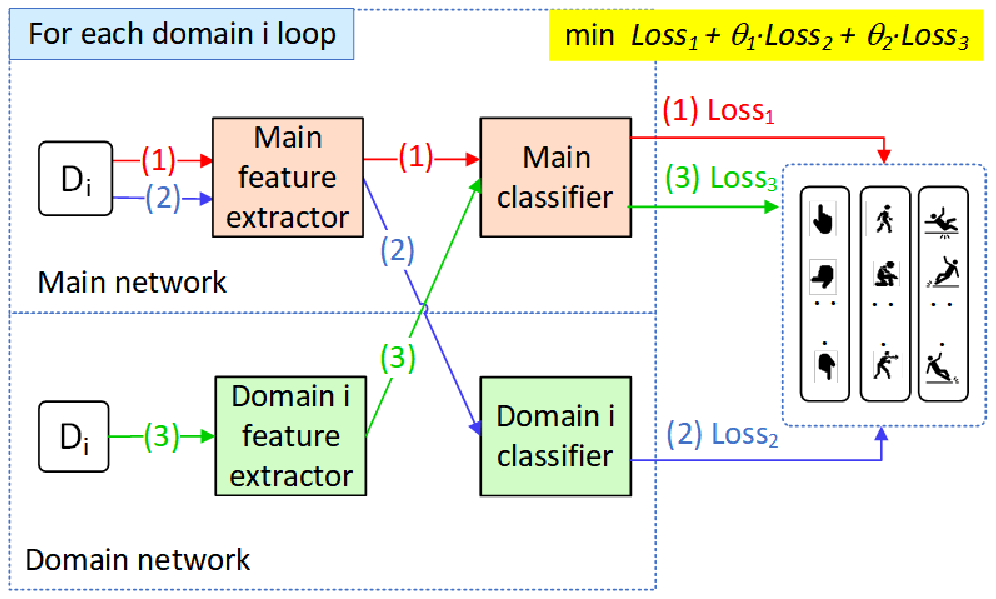}}
\caption{Episodic training for domain independent feature extraction.}
\label{FigEpisodic}
\end{figure*}

The domain networks are first trained. For each domain $i$, its dataset $D_i^s$ is input to the corresponding domain network, so that the domain network focuses on feature extraction and classification within domain $i$, as illustrated in Fig.~\ref{FigEpisodic:b}. During the training of the domain network $i$, the training samples, denoted as $x_{ij}^{s} \in D_i^s$, are input to the domain feature extractor. The features are extracted as:
\begin{equation}
    f_{ij}^{s} = Extractor_i(x_{ij}^{s})
\end{equation}
where $Extractor_i(\cdot)$ represents the feature extractor of the domain network $i$. The features are passed to the corresponding domain classifier to obtain the classification results as:
\begin{equation}
    \hat{y}_{ij}^{s} = Classifier_i(f_{ij}^{s})
\end{equation}
where $Classifier_i(\cdot)$ represents the classifier of the domain network $i$. The loss of the domain network $i$ is defined as:
\begin{equation}
    {\rm min} \quad L_i = \frac{1}{n_i}\sum_{j=1}^{n_i}CR(\hat{y}_{ij}^{s}, y_{ij}^{s})
\end{equation}
where $n_i$ is the number of samples in $D_i^s$, $CR(\cdot)$ represents cross entropy. All the domain networks are trained.

\subsubsection{Training of the main network}
The training of the main network requires the participation of all the source domain data. The main network is optimized through episodic training with all the domain networks, enabling it to extract domain independent features. To ensure the robustness of the main feature extractor, it is required that the features extracted by the main feature extractor can be recognized by each domain classifier. Likewise, for a robust main classifier, it is required that the main classifier can recognize the features extracted by each domain feature extractor. To meet these requirements, the training process of the main network is as follows~\cite{LiD:2019}. As depicted in Fig.~\ref{FigEpisodic:c}, repeat the following steps (1)--(3) for each domain $i$:

(1) Input the samples $(x_{ij}^{s},y_{ij}^{s}) \in D_{i}^{s}$ of the source domain $i$ to the main network, as the flow (1) shows in Fig.~\ref{FigEpisodic:c}. Assume $MExtractor$ stands for the main feature extractor and $MClassifier$ stands for the main classifier. The loss is defined as:
\begin{equation}
    Loss_1 = CR(MClassifier(MExtractor(x_{ij}^{s})),y_{ij}^{s})
\end{equation}
This step is to ensure that the features extracted by the main feature extractor can be recognized by the main classifier.

(2) Input the samples $(x_{ij}^{s},y_{ij}^{s}) \in D_{i}^{s}$ of the source domain $i$ to the main feature extractor. The extracted features are then input to the classifier $Classifier_i$ of domain $i$ for classification, as the flow (2) shows in Fig.~\ref{FigEpisodic:c}. The loss is defined as:
\begin{equation}
    Loss_2 = CR(Classifier_i(MExtractor(x_{ij}^{s})),y_{ij}^{s})
\end{equation}
This step is to ensure that the features extracted by the main feature extractor can be recognized by the domain classifier, i.e., the main feature extractor is robust.

(3) Input the samples $(x_{ij}^{s},y_{ij}^{s}) \in D_{i}^{s}$ of the source domain $i$ to the feature extractor $Extractor_i$ of the domain $i$. The extracted features are then input to the main classifier for classification, as the flow (3) shows in Fig.~\ref{FigEpisodic:c}. The loss is defined as:
\begin{equation}
    Loss_3 = CR(MClassifier(Extractor_i(x_{ij}^{s})),y_{ij}^{s})
\end{equation}
This step is to ensure that the main classifier can recognize the features extracted by the domain feature extractor, i.e., the main classifier is robust.

During the training of the main network, the parameters of the domain networks are fixed, only the parameters of the main network are optimized. The overall loss of the main network is defined as:
\begin{equation} 
\footnotesize
		 {\rm min} \ L_{main} = \frac{1}{N}\sum_{i=1}^{N}\left(\frac{1}{n_i}\sum_{j=1}^{n_i}(Loss_1 + \theta_{1} Loss_2 + \theta_{2} Loss_3)\right)
\end{equation}
Each source domain trains with the main network in turn. After training, the main network acquires the generalization ability across domains.

\begin{figure*}
\begin{center}
		\subfigure[Episodic step 1]{
		\label{FigEpisodicEffect:a}
		\includegraphics[height=2.7cm]{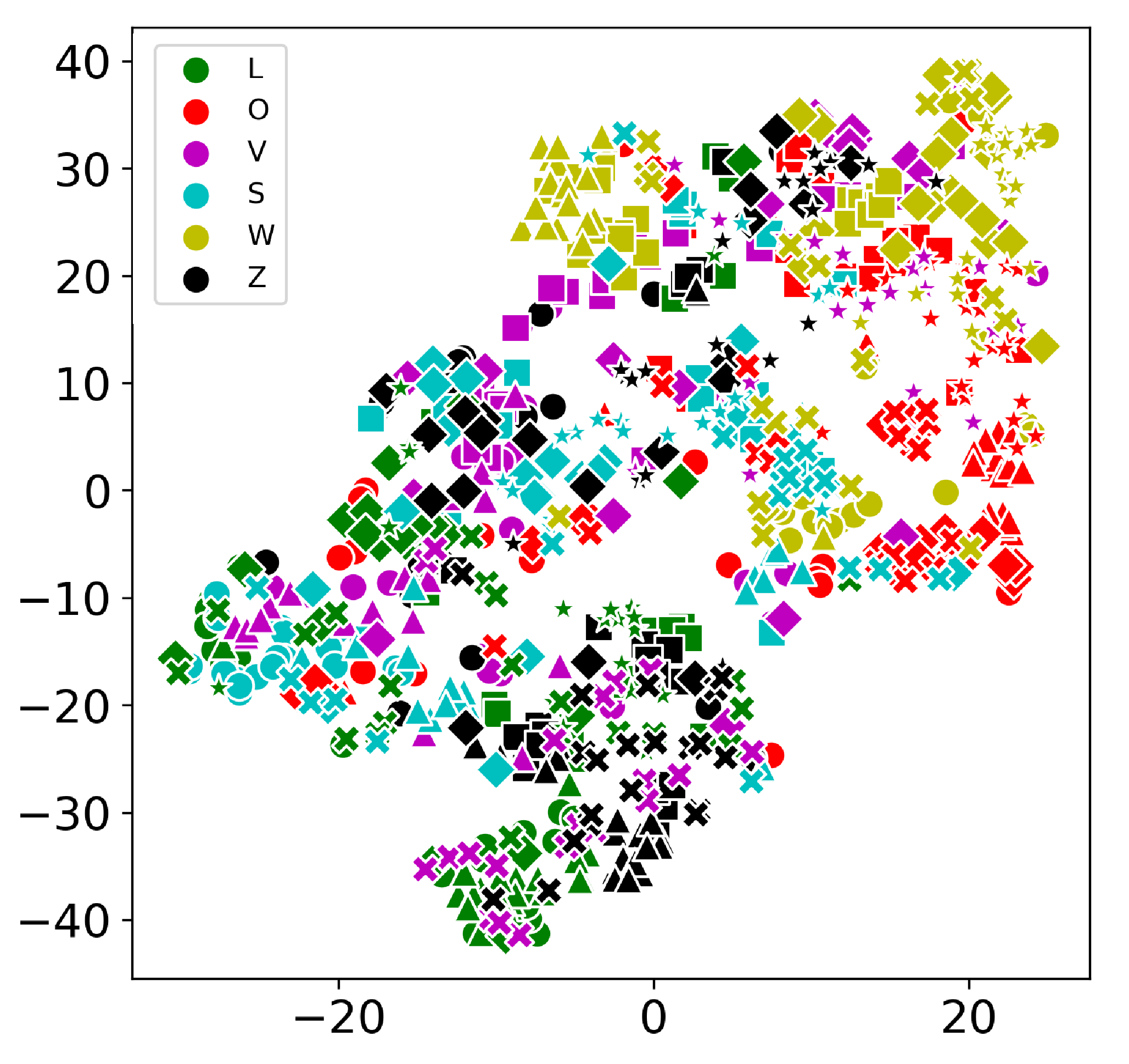}}
		\hspace{0.2cm}
		\subfigure[Episodic step 2]{
		\label{FigEpisodicEffect:b}
		\includegraphics[height=2.7cm]{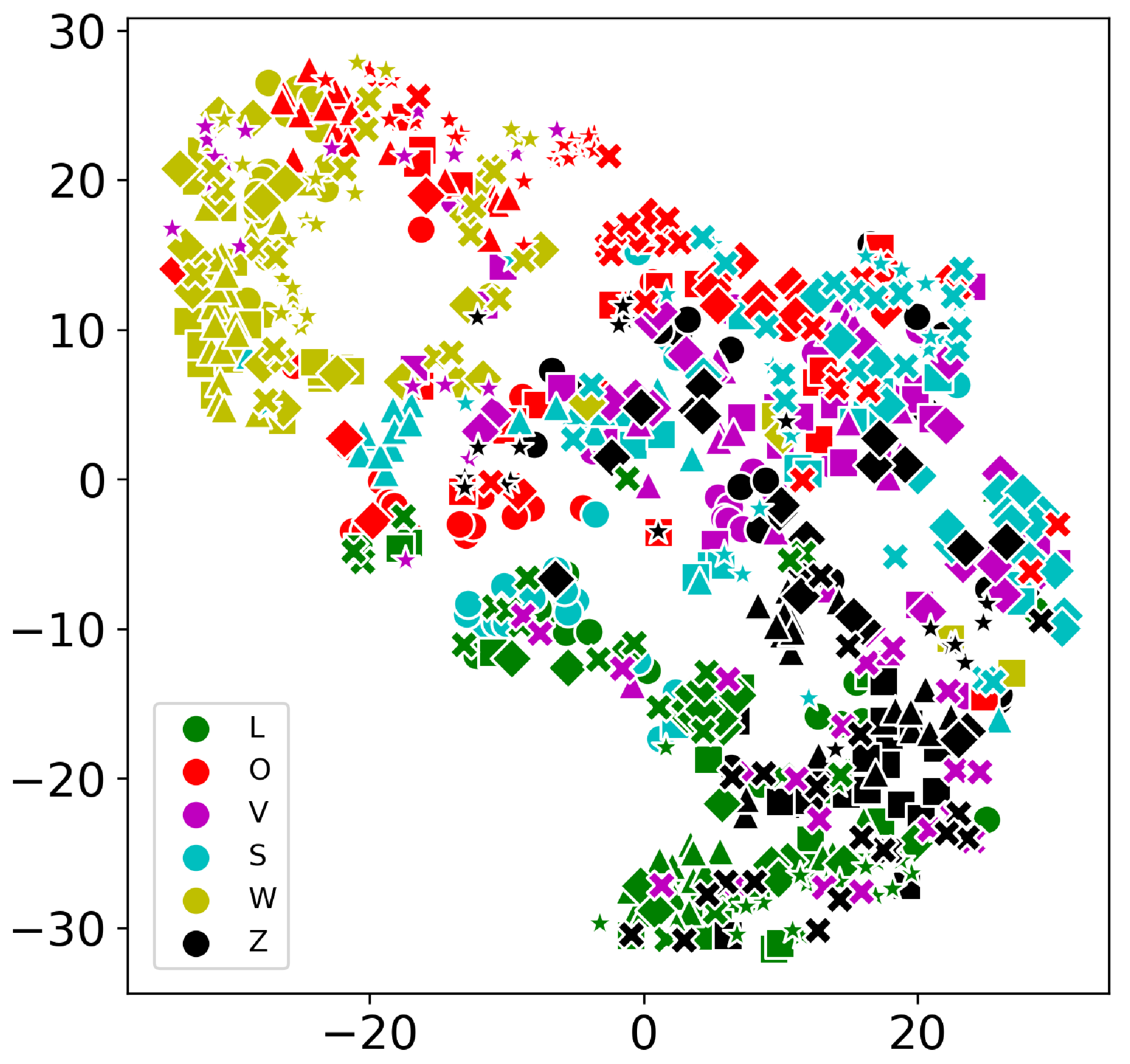}}
		\hspace{0.2cm}
		\subfigure[Episodic step 3]{
		\label{FigEpisodicEffect:c}
		\includegraphics[height=2.7cm]{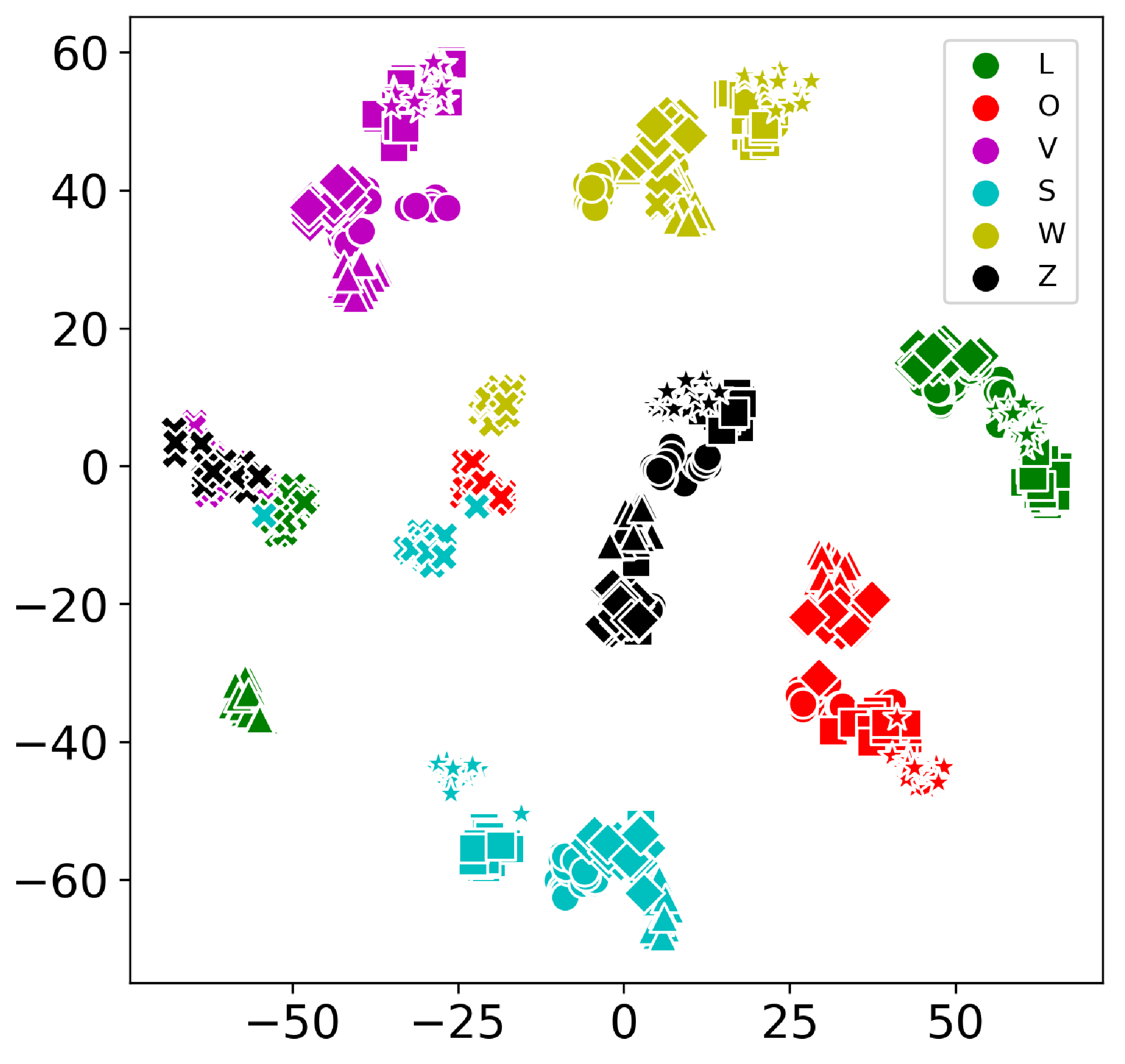}}
		\hspace{0.2cm}
		\subfigure[Episodic step 4]{
		\label{FigEpisodicEffect:d}
		\includegraphics[height=2.7cm]{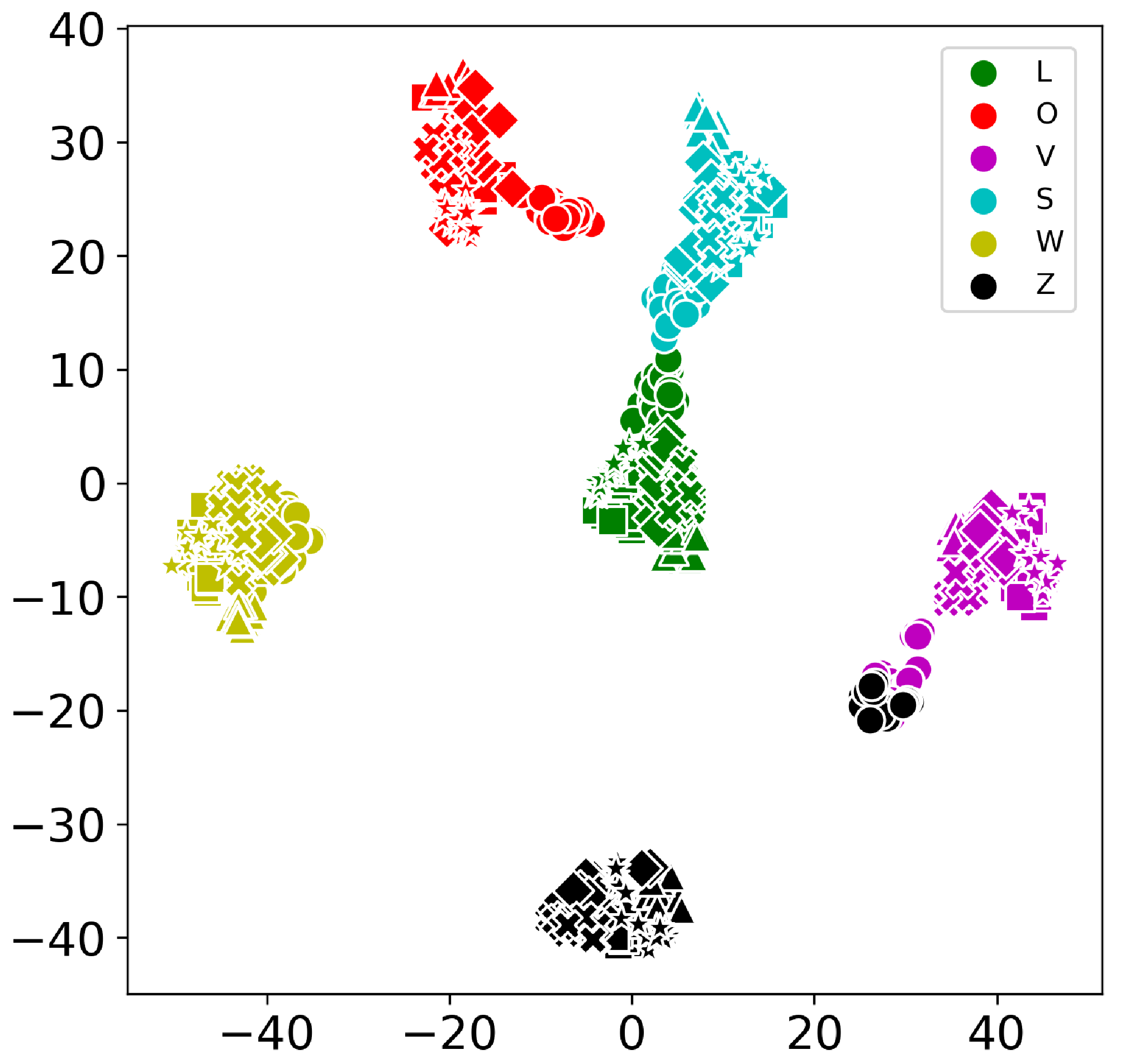}}
		\hspace{0.2cm}
		\subfigure[Episodic step 5]{
		\label{FigEpisodicEffect:e}
		\includegraphics[height=2.7cm]{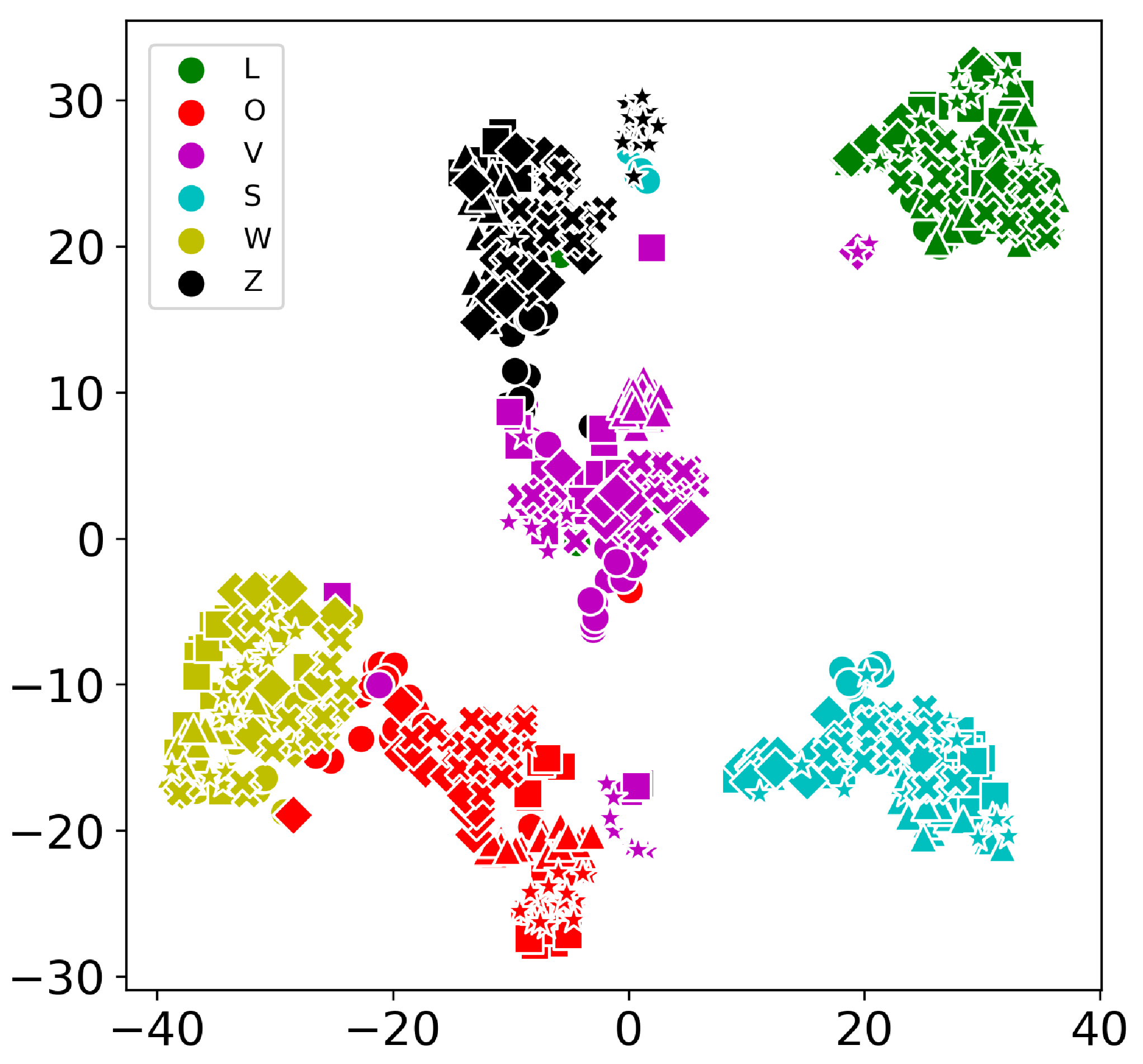}}
\caption{Process and effect of episodic training on WiFi gesture data (5 training users, 1 new user).}
\label{FigEpisodicEffect}
\end{center}
\end{figure*}

To demonstrate the process and effect of episodic training, we plot the t-SNE~\cite{tSNE:2008} visualization of WiFi gesture data of 6 users after each episodic step, as shown in Fig.~\ref{FigEpisodicEffect}. The colors represent the gesture types, and the shapes represent the users. We took 5 users as the source domains and the rest 1 user as the new domain (i.e. unseen target domain). Hence the training process went through 5 episodic steps. Each episodic step trained the main network with a domain network (i.e. a user). From the t-SNE plots we can see that the gesture data aggregate to gesture types step by step. We also made gesture recognition tests on the new user after each episodic step. From the 1st episodic step to the 5th episodic step, the recognition accuracy increased from 24.2\%, 33.3\%, 60.8\%, 75.0\%, to 83.3\%, indicating that the generalization ability of the framework grew with the training.

\subsubsection{Classification}
By augmenting the training set and conducting episodic training, the main network acquires the ability to extract domain independent features and sense across domains. When deploying the model in a new/unseen domain, the process of recognition is straightforward. As illustrated in Fig.~\ref{FigEpisodic:a}, the data, denoted as $x_{i}^{t}$, from the new domain is input to the main feature extractor, which extracts only the sensing-relevant features. These features are fed into the main classifier to obtain the sensing results as:
\begin{equation}
    \hat{y}_{i}^{t} = MClassifier(MExtractor(x_{i}^{t}))
\end{equation}

\section{Evaluations on WiFi gesture recognition}
\label{SecWiFi}

\subsection{WiFi signals and preprocessing}
We exploited amplitude, phase, and Doppler spectrogram from the CSI data to realize gesture recognition. The amplitude and phase depict the received power and phase on each subcarrier, which can be retrieved from certain commodity network interface cards with CSI tools~\cite{Halperin:2010, XieY:2015}. The received CSI is a matrix ${\mathrm{H}}=\left( {H}_{ij} \right)_{N_{tx}\times N_{rx}}$, where $N_{tx}$ is the number of transmitting antennas, $N_{rx}$ is the number of receiving antennas, ${H}_{ij}$ is the CSI of the channel formed by transmitting antenna $i$ and receiving antenna $j$, containing $N_s$ subcarriers, expressed as ${H}_{ij}=(h_1,h_2,\cdots,h_{N_s})$. The $k$-th subcarrier in ${H}_{ij}$ can be expressed as $h_k=|h_k|e^{j\angle h_k}$, where $|h_k|$ is the amplitude and $\angle h_k$ is the phase.  

Human bodies and movements may alter the amplitude and the phase of the received signals. Human motions may cause changes in the lengths of reflection paths, resulting in Doppler effect. We exploit the amplitude and the phase of the subcarriers as well as Doppler spectrograms to recognize gestures. The raw amplitude data are denoised by the median filter, and the raw phase data are sanitized by a linear transformation method~\cite{QianK:2014}. The Doppler spectrogram is generated through antenna selection, bandpass filtering, dimension reduction and Short Time Fourier Transform (STFT) on amplitude time-series~\cite{QianK:2017}. The amplitude, the phase and the spectrogram are originated from the same sensing medium. They represent different but correlated information and depict different aspects of the same signal. After preprocessing, they form respective samples for the same gestures, hence we regard them as three modalities and fuse them for gesture recognition. 

\subsection{System design} 
\begin{figure}
\begin{center}
		\includegraphics[width=7cm]{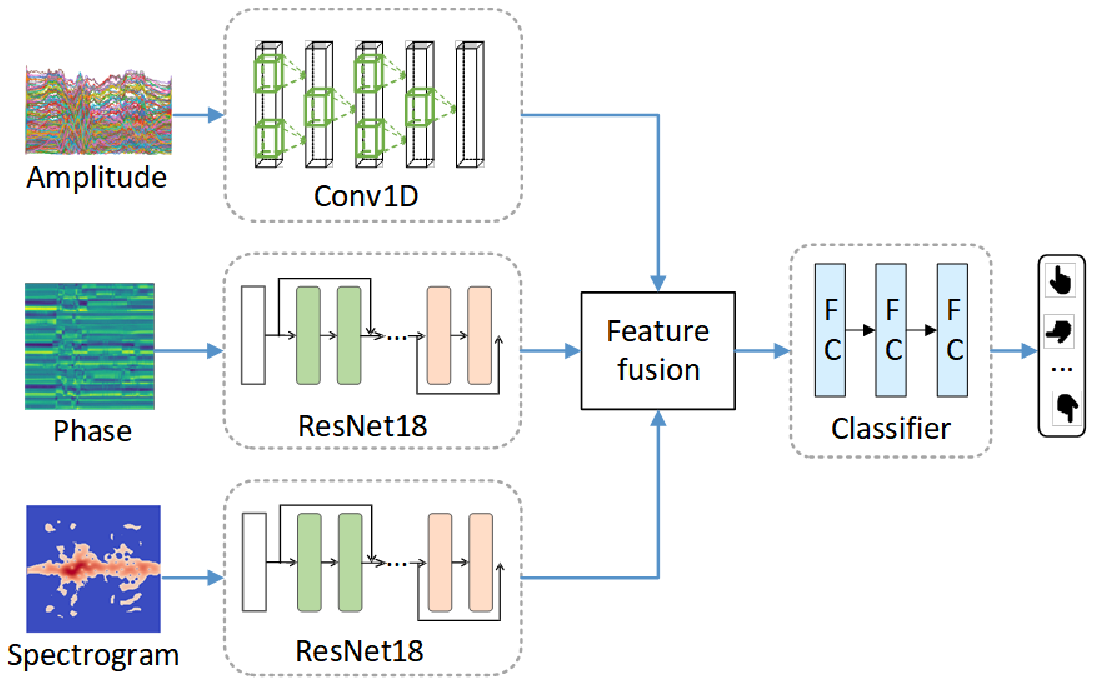}
\caption{The feature extractor and classifier of WiFi gesture recognition.}
\label{WiFiSystem}
\end{center}
\end{figure}

The system is based on the DGSense framework, comprising data acquisition and preprocessing, virtual data generation, domain independent feature extraction and classification. As there are three modalities, we employ the cross-modal virtual data generator to enhance the diversity of the training set. The virtual data generator takes the amplitude as the base modality, establishes the mapping between the amplitude and the other modalities during training, and generates the virtual data of the three modalities from the real amplitude. This ensures consistency between the virtual modalities of the same gesture. Episodic training is applied to extract domain independent features. As there are three modalities, we design a composite feature extractor, comprising an amplitude feature extractor, a phase feature extractor and a spectrogram feature extractor. As illustrated in Fig.~\ref{WiFiSystem}, the amplitude feature extractor is 1DCNN to extract the temporal features, the phase feature extractor and the spectrogram feature extractor adopt ResNet18 to extract the spatial features. The features from the three modalities are fused and fed to the classifier for gesture recognition. Assuming the sample is denoted as $x=(x^{am},x^{ph},x^{sp})$, the feature fusion and classification can be formulated as:
\begin{equation}
\small
\begin{split}
f &= \alpha_1 AmExt(x^{am}) + \alpha_2 PhExt(x^{ph}) + \alpha_3 SpExt(x^{sp})	\\
y &= Classifier(f) 
\end{split}
\end{equation}

\subsection{Evaluations}
\begin{figure*}
\begin{center}
\subfigure[Lounge (R1)]{
\includegraphics[height=2cm]{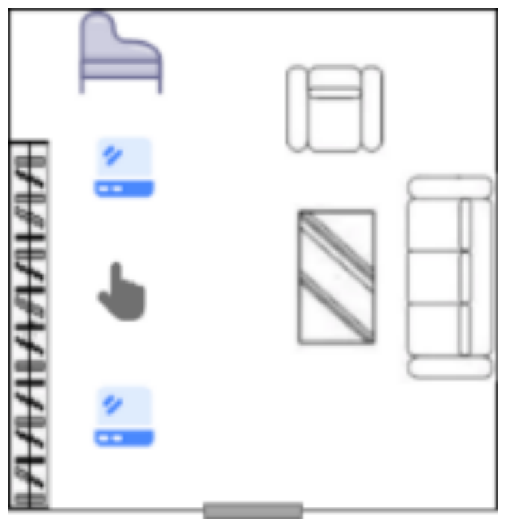}}
\hspace{0.5cm}
\subfigure[Laboratory (R2)]{
\includegraphics[height=2cm]{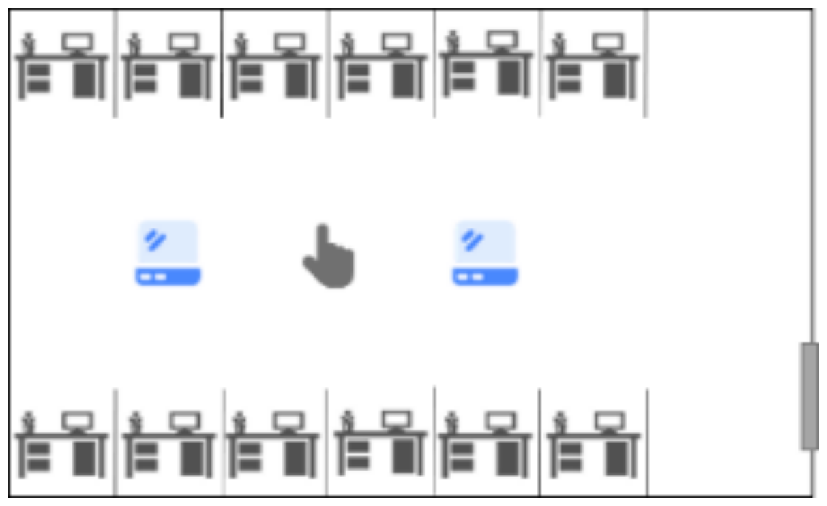}}
\hspace{0.5cm}
\subfigure[Large meeting room (R3)]{
\includegraphics[height=2cm]{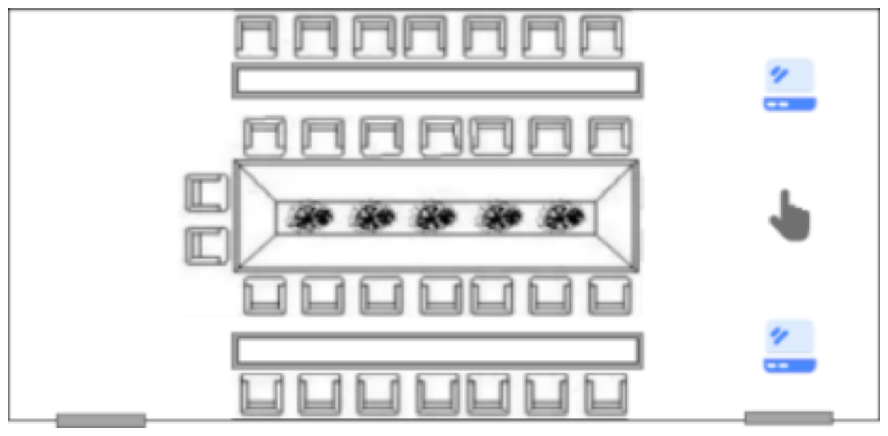}}
\hspace{0.2cm}
\subfigure[Small meeting room (R4)]{
\includegraphics[height=2cm]{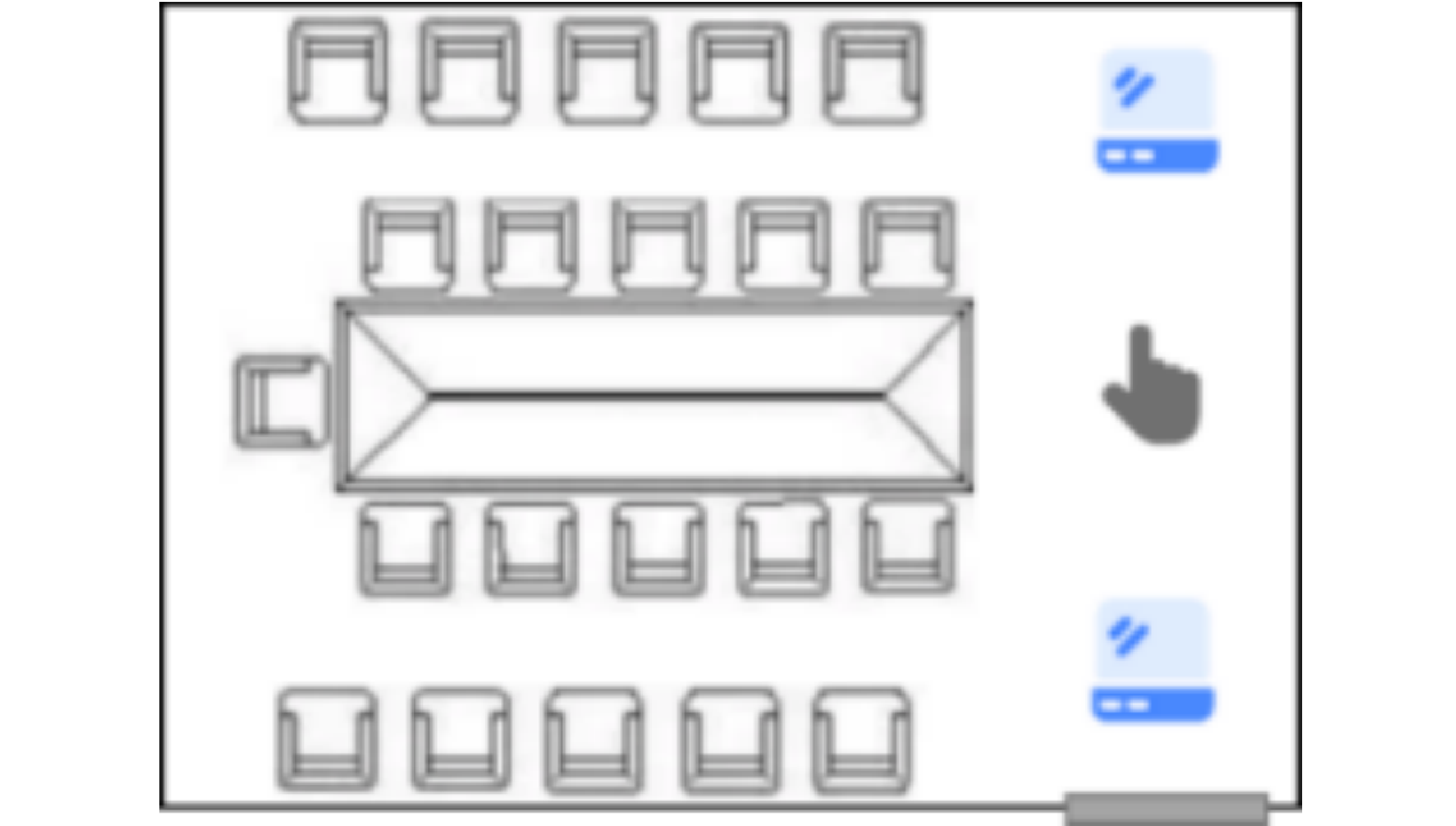}}
\caption{WiFi experimental setup.}
\label{WiFiExperimentSetup}
\end{center}
\end{figure*}

We deployed two laptops equipped with Intel 5300 wireless adapters, one as the transmitter and the other as the receiver. Each transceiver had 3 antennas, forming 9 links. Each link contained 30 subcarrier groups, resulting in 270 dimensions. The sampling frequency was 1000Hz to capture the Doppler effect. The gesture data were collected in 4 rooms, as shown in Fig.~\ref{WiFiExperimentSetup}, with the sizes of 5m$\times$5m, 6m$\times$8m, 11.4m$\times$6.8m, and 7.6m$\times$5.8m. The distance between the transmitter and the receiver was about 2m. During the experiments, 6 volunteers traced the letters \{L, O, V, S, W, Z\} with hand gestures in each room. Each gesture was repeated 20 times, resulting in a total of 720 gesture samples in each room. 

\begin{figure*}
\begin{center}
\subfigure[New user (5 source 1 new)]{
\label{FigWiFiGesture:a}
\raisebox{0.5cm}{\includegraphics[height=2.5cm]{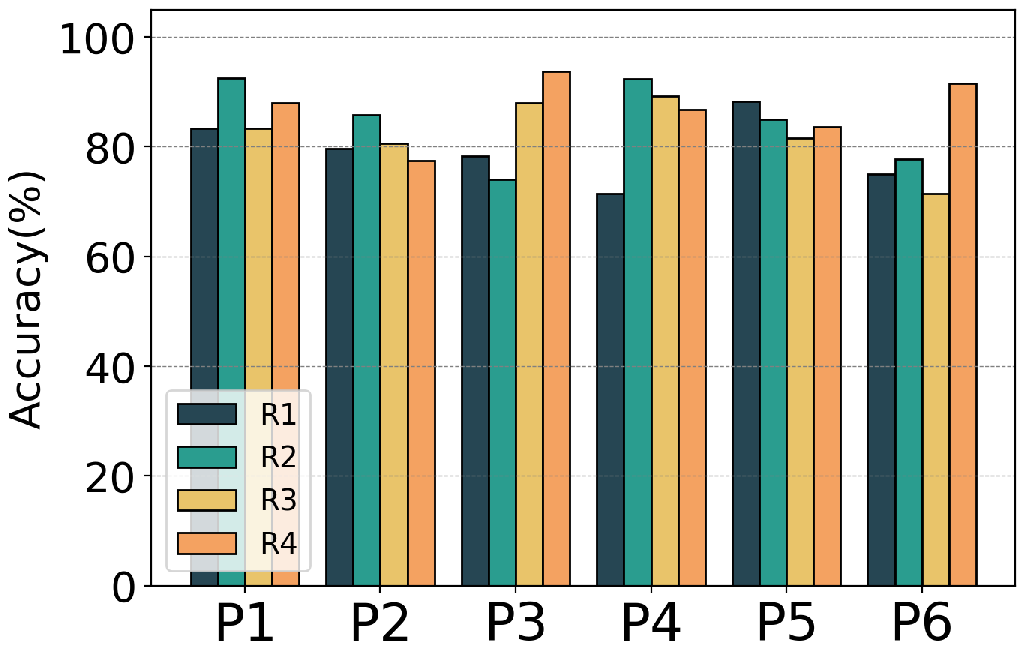}}}
\subfigure[New user (4 source 2 new)]{
\label{FigWiFiGesture:b}
\raisebox{0.5cm}{\includegraphics[height=2.5cm]{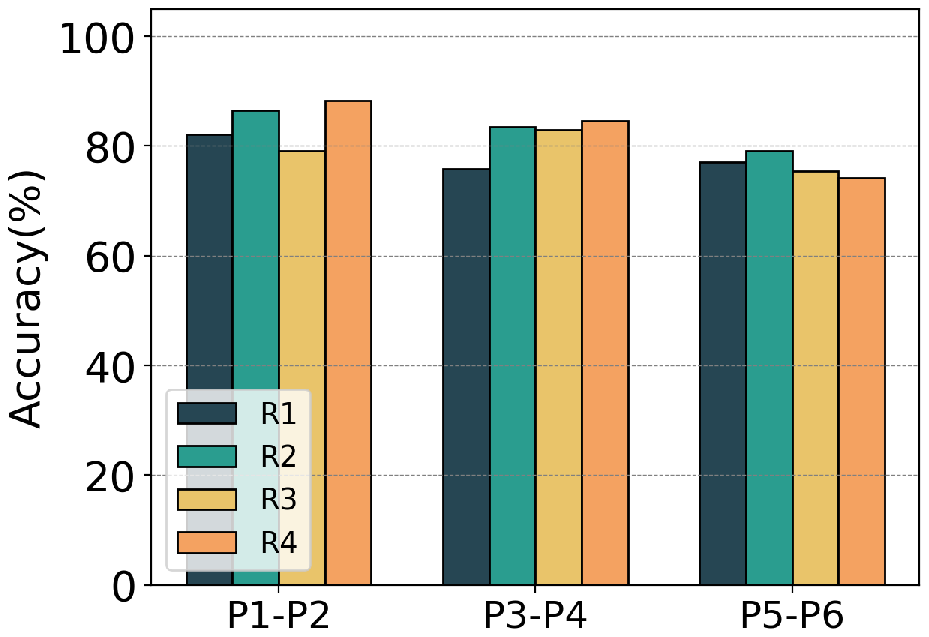}}}
\subfigure[New room]{
\label{FigWiFiGesture:c}
\raisebox{0.5cm}{\includegraphics[height=2.5cm]{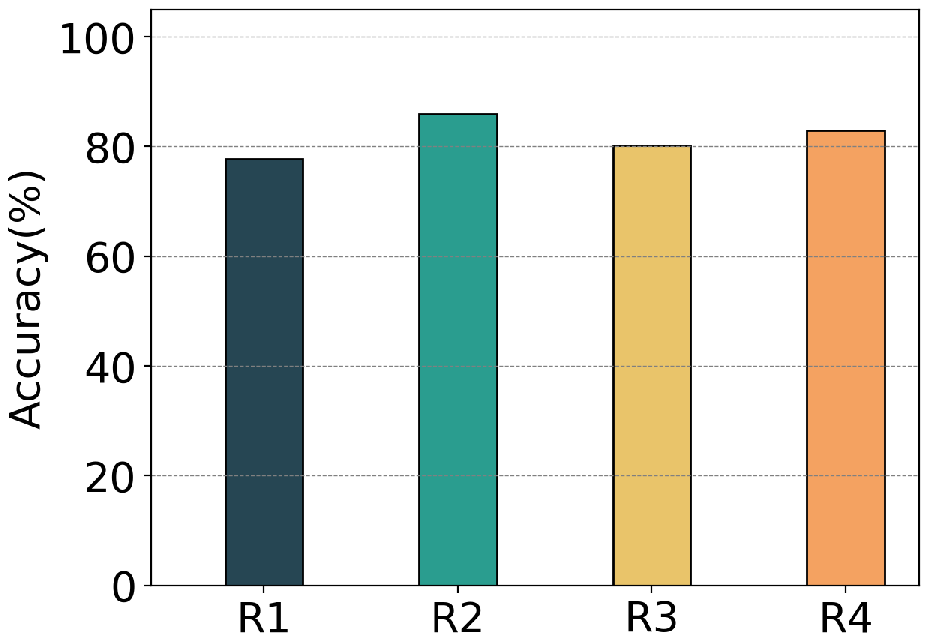}}}
\subfigure[New user and new room]{
\label{FigWiFiGesture:d}
\includegraphics[height=3.0cm]{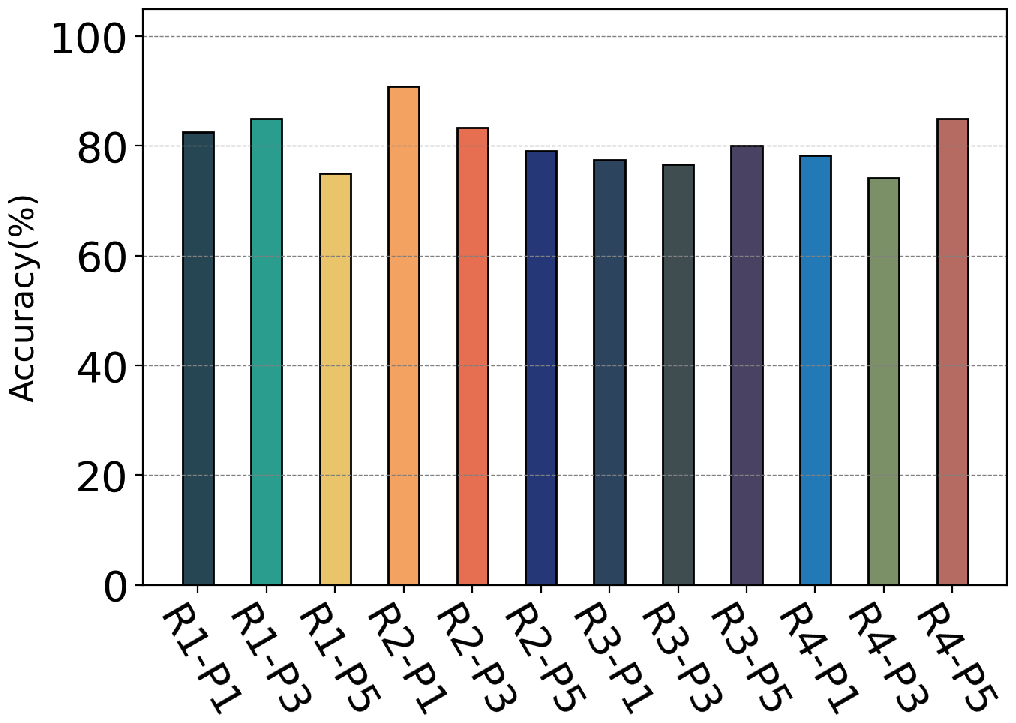}}
\caption{WiFi gesture recognition in new domains.}
\label{FigWiFiGesture}
\end{center}
\end{figure*}

\subsubsection{In-domain accuracy}
We first evaluated the in-domain accuracy of the WiFi gesture recognition system, involving 6 gestures by 6 volunteers in 4 rooms. Each gesture was repeated 20 times by each volunteer in each room. We conducted 5-fold cross-validation on the dataset and achieved the average accuracy of 97.8\% for familiar users in seen rooms.

\subsubsection{Quality of virtual data}
We augmented the training set by cross-modal virtual data generation. The number of virtual samples equaled the number of real samples. We conducted experiments to verify the quality of the virtual data, using the method proposed in DANGR~\cite{HanZ:2020}. We trained the feature extractor and classifier with the real data and tested on the virtual data. The classification accuracy achieved 97.0\%, close to the in-domain accuracy of 97.8\%, verifying that the virtual data followed the same distribution with the real data. We then trained the feature extractor and classifier with the virtual data and tested on the real data. The classification accuracy achieved 94.4\%, showing that the virtual data brought adequate diversity.

\subsubsection{New user}
To evaluate the generalization of the method on new users, we conducted two groups of experiments involving 6 volunteers in 4 rooms. In the first experiment, we designated 5 volunteers as the source domains and tested on the left 1 volunteer as the new domain. The testings were conducted on each volunteer in turn in each room. The results are shown in Fig.~\ref{FigWiFiGesture:a}, where P1--P6 represent the volunteers, R1--R4 represent the rooms. The columns show the recognition accuracy of each volunteer as the new user in each room. The average gesture recognition accuracy reached 83.3\% for new users. In the second experiment, we selected 2 volunteers (P1-P2, P3-P4, and P5-P6) as the new domains while the other 4 volunteers acted as the source domains. The testings were conducted on each volunteer group in turn in each room. The results are shown in Fig.~\ref{FigWiFiGesture:b}, in which the columns show the recognition accuracy of each volunteer group as new users in each room. The average gesture recognition accuracy reached 80.6\% for new users. By leveraging the proposed DGSense framework, the recognition accuracy improved from around 25\% (without virtual data generation and domain generalization) to around 80\%.

\subsubsection{New room}
To evaluate the generalization of the method in new rooms, we conducted experiments in 4 different rooms involving 6 volunteers. We took one room as the new domain and utilized the other 3 rooms as the source domains. The testings were conducted on each room as the new room in turn. The results are illustrated in Fig.~\ref{FigWiFiGesture:c}, where the columns display the recognition accuracy in each room when treated as the new domain. The average gesture recognition accuracy achieved 81.7\% in new rooms. By leveraging the proposed DGSense framework, the recognition accuracy improved from around 20\% (without virtual data generation and domain generalization) to around 80\%.

\subsubsection{New user and new room}
We also conducted experiments to evaluate the generalization of the method on new users in new rooms. We designated 1 user and 1 room as the new domain, while considering the other 5 users in the other 3 rooms as the source domains. The user-room combinations and the gesture recognition results are illustrated in Fig.~\ref{FigWiFiGesture:d}, where the columns indicate the recognition accuracy of the user-room combinations as the new domains. The average gesture recognition accuracy achieved 80.6\% for new users in new rooms. 

\subsubsection{Comparison with existing works}
\begin{figure}
\begin{center}
\subfigure[New user]{
\label{FigWiFiComparasion:a}
\includegraphics[height=2.5cm]{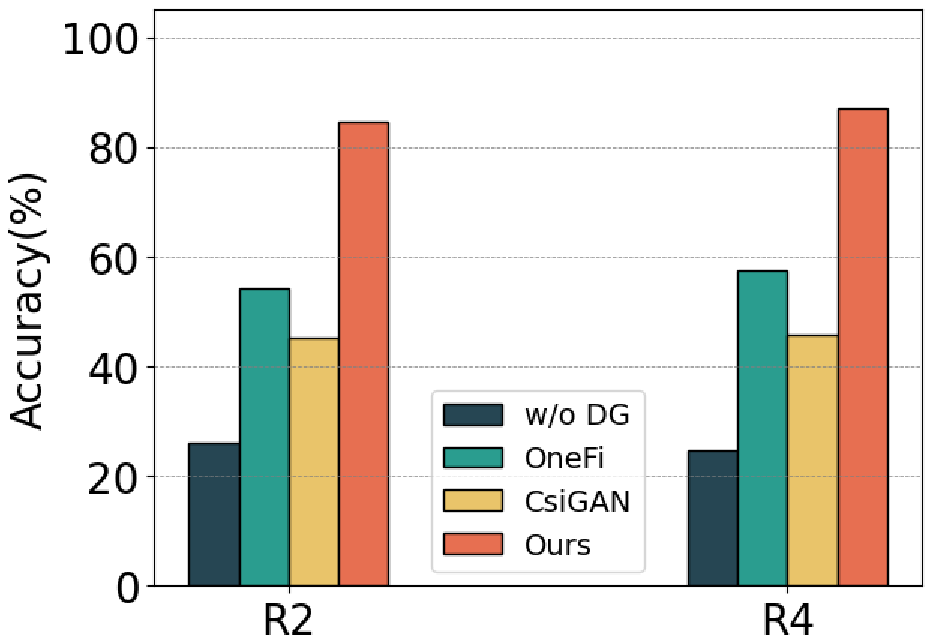}}
\hspace{0.5cm}
\subfigure[New room]{
\label{FigWiFiComparasion:b}
\includegraphics[height=2.5cm]{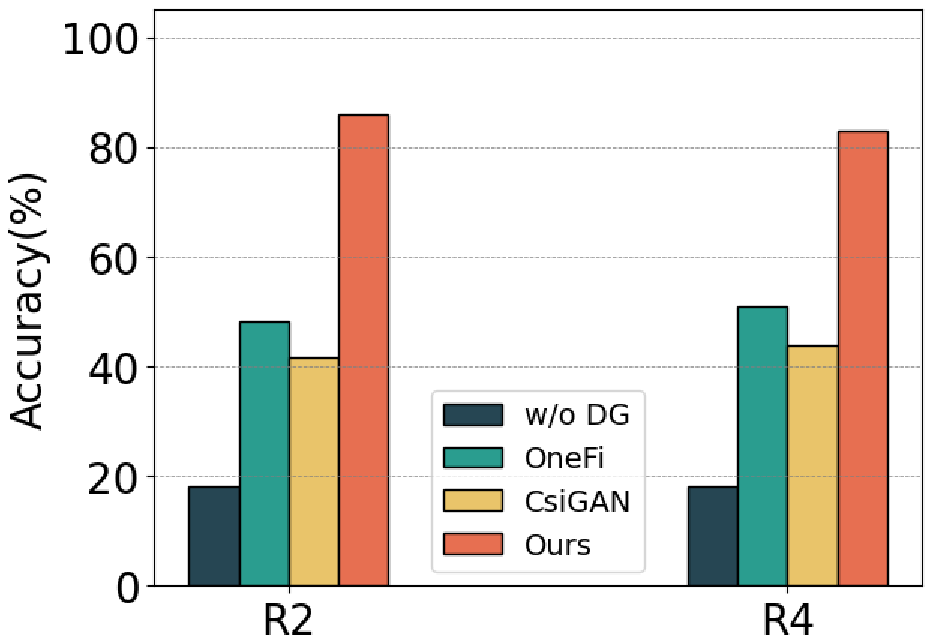}}
\caption{Comparison of WiFi gesture recognition with existing works.}
\label{FigWiFiComparasion}
\end{center}
\end{figure}

To validate the effectiveness of the proposed method, we compared with the methods of \textit{w/o DG}, \textit{OneFi}~\cite{XiaoR:2021} and \textit{CsiGAN}~\cite{XiaoCJ:2019}. The \textit{w/o DG} method excludes virtual data generation and episodic training. \textit{OneFi}~\cite{XiaoR:2021} introduces a one-shot recognition framework based on Doppler spectrogram and virtual data generation. \textit{CsiGAN}~\cite{XiaoCJ:2019} leverages semi-supervised GAN to generate virtual data and trains a robust classifier using some target domain data. We conducted two groups of comparative experiments. The first group compared their performance on new users. To this end, we took 1 volunteer as the new domain in turn and the other 5 volunteers as the source domains, in the laboratory (referred to as R2) and the small meeting room (referred to as R4). Each method utilized only the source domain data to train the model and tested in the new domain. The comparison results are shown in Fig.~\ref{FigWiFiComparasion:a}, in which the columns represent the average recognition accuracy of each method on new users. The second group compared their performance in new rooms. We took the laboratory (R2) and the small meeting room (R4) as the new domain respectively and considered the other 3 rooms as the source domains, with the participation of 6 volunteers. Each method utilized only the source domain data for model training and tested in the new room. The results are shown in Fig.~\ref{FigWiFiComparasion:b}, in which the columns represent the average recognition accuracy of each method in new rooms. Across all the testings, our method consistently achieved the highest accuracy. The reason is that our approach is a domain generalization method that combines virtual data generation and domain independent feature extraction. In contrast, \textit{OneFi} and \textit{CsiGAN} adopt domain adaptation techniques that rely on some target domain data, resulting in limited generalization ability in the absence of such data. They all outperformed the method without domain generalization by a large margin.  

\subsubsection{Computation costs}
The system ran on a desktop equipped with a GPU of Colorful RTX 3090 and a CPU of Intel i5-12600K. For the training of 2160 samples (involving 3 rooms, 6 users, 6 gestures), the system spent 117s for data preprocessing, 2328s to train the virtual data generator, 1s to generate 2160 virtual samples, and 22243s for episodic training in 3 source domains. In total, it took 24689s (6.86h) for the whole training process. To recognize a gesture, the system spent 41.2ms for data preprocessing, 1.5ms to extract the features and classify the gesture. In total, it took 42.7ms to recognize a gesture. Although the training took a few hours, the recognition was in real-time.

\section{Evaluations on mmWave activity recognition}
\label{SecMmWave}

\subsection{mmWave signals and preprocessing}
\begin{figure}
\centerline{\includegraphics[width=7cm]{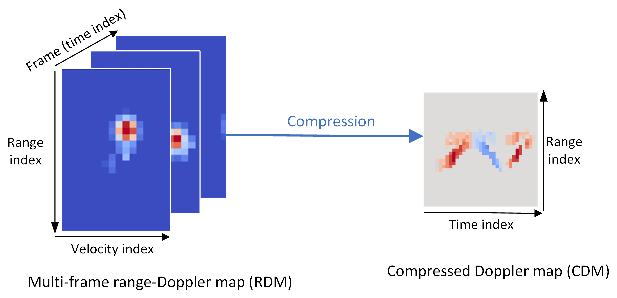}}
\caption{The mmWave data and preprocessing.}
\label{FigMmWaveData}
\end{figure}

We deploy the TI IWR1443 boost radar board to collect mmWave data, which operates in the frequency range of 76--81GHz and has a bandwidth of 4GHz for FMCW signals. The mmWave radar contains 3 transmitting antennas and 4 receiving antennas. We retrieve the range-Doppler data from the radar to recognize activities. The data are time-series of range-Doppler maps, describing the range and velocity of the target relative to the radar over time, which can be expressed as ${RDM} = ({RDM}_{1}, {RDM}_{2}, \cdots, {RDM}_{t})$. ${RDM}_{i}$ is the $i$-th frame in the time-series, which is a range-Doppler map and can be expressed as a matrix ${RDM}_{i}=(p^i_{rv})$.
in which $r$ represents the range index, $v$ represents the velocity index, and the values represent the power of the signal. Different activities cause different patterns in the multi-frame range-Doppler maps, which can be used to deduce the activities. To reduce the computing complexity, we compress the multi-frame range-Doppler maps from 3D to 2D, by compressing the velocity dimension and using the velocity with the maximal probability, as illustrated in Fig.~\ref{FigMmWaveData}. The compressed 2D Doppler map is derived as:
\begin{equation} 
{CDM} = (\mathop{\arg\max}_v{{RDM}_{1}},\cdots,\mathop{\arg\max}_v{{RDM}_{t}})
\end{equation}
In the compressed Doppler map ${CDM}$, one dimension represents the time index and the other represents the range index, and the values represent the velocities of the target, reflecting the change of velocity over time in terms of distance. We use different colors to indicate the velocity values. The red color represents the positive velocity away from the radar, while the blue color represents the negative velocity approaching the radar. We take the images of the compressed Doppler maps as the input to the sensing model. 

\subsection{System design}

\begin{figure}
\begin{center}
\includegraphics[width=8cm]{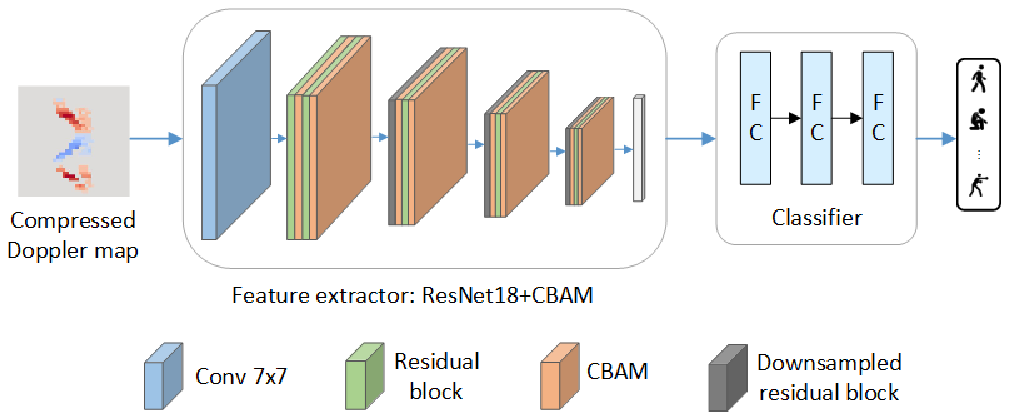}
\caption{The feature extractor and classifier of mmWave activity recognition.}
\label{mmWaveSystem}
\end{center}
\end{figure}

The system is based on the DGSense framework, consisting of data collection and preprocessing, virtual data generation, domain independent feature extraction and classification. Given that the samples take the form of images, we employ a single-modal virtual data generator. The virtual samples are generated by adding noise to the intermediate features and decoding them. Feature extraction and classification involves a main network and several domain networks, exploiting episodic training to extract domain independent activity features. The main network and the domain networks share the same structure, as shown in Fig.~\ref{mmWaveSystem}. The feature extractor is based on ResNet18 and enhanced by CBAM within each residual block, so that the feature extractor emphasizes the activity region. The inclusion of CBAM brings together spatial attention and channel attention mechanisms~\cite{Woo:2018}, enabling the differentiation of activities and enhancing the distinction between similar activities. The classifier is a 3-layer fully connected neural network.  

\subsection{Evaluations}
\begin{figure*}
\begin{center}
\subfigure[The testbed]{
\label{MmWaveExperimentSetup:a}
\raisebox{0.25cm}{\includegraphics[width=3.5cm]{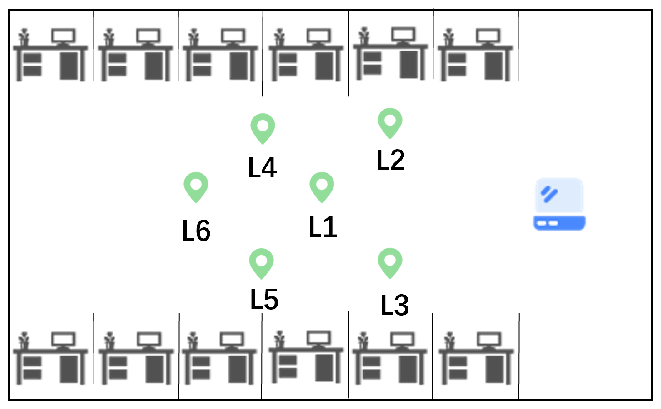}}}
\hspace{0.5cm}
\subfigure[New user]{
\label{MmwaveActivityResults:a}
\includegraphics[width=3.5cm]{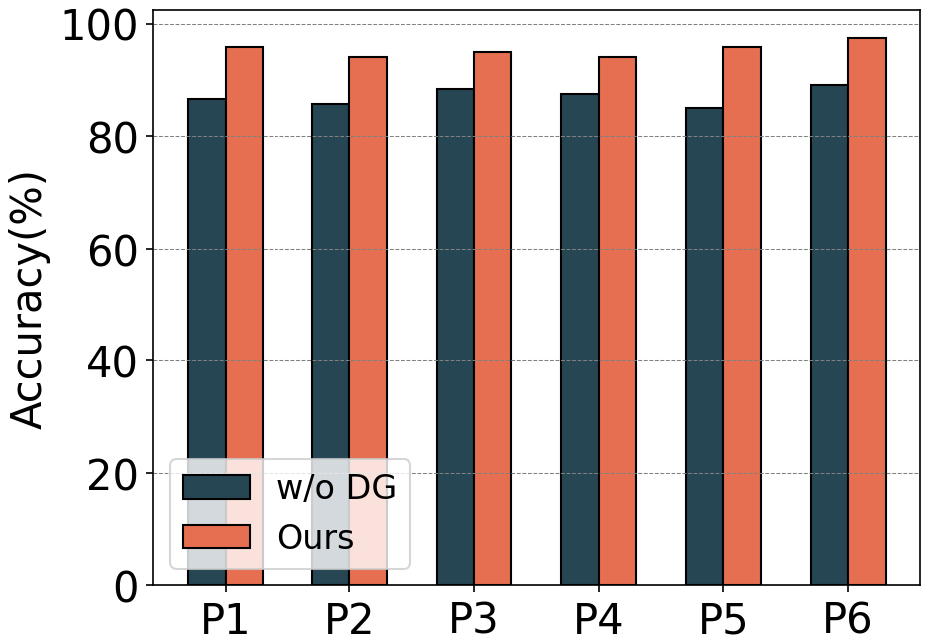}}
\hspace{0.5cm}
\subfigure[New location]{
\label{MmwaveActivityResults:b}
\includegraphics[width=3.5cm]{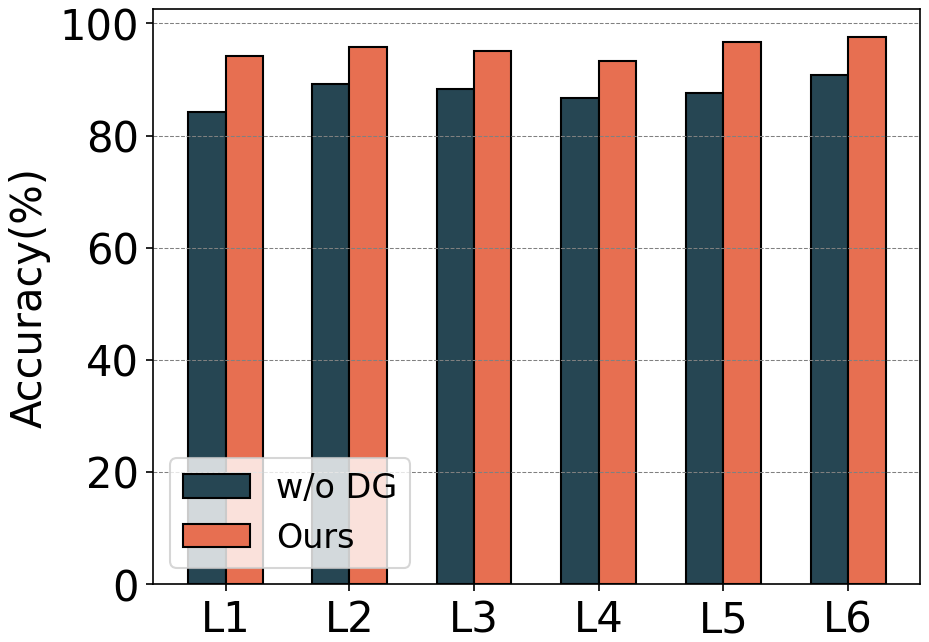}}
\caption{Experimental setup and mmWave activity recognition in new domains.}
\label{MmwaveActivityResults}
\end{center}
\end{figure*}

The experiments were carried out in a laboratory, as shown in Fig.~\ref{MmWaveExperimentSetup:a}. The laboratory had the size of 6m$\times$8m and was cluttered with computers on both sides. The activities were performed at 6 locations (referred to as L1--L6) in the laboratory, as marked on Fig.~\ref{MmWaveExperimentSetup:a}, with location 1 being 2m away from the mmWave radar and the adjacent locations being 1m apart. Six volunteers (referred to as P1--P6) participated in the experiments. The volunteers performed 6 activities: squatting down, standing up, jumping, standing still, boxing, and walking. Each activity was repeated 20 times by each person at each location. 

\begin{figure*}
\begin{center}
\subfigure[Learning models]{
\label{FigMmwaveComparison:a}
\includegraphics[height=2.8cm]{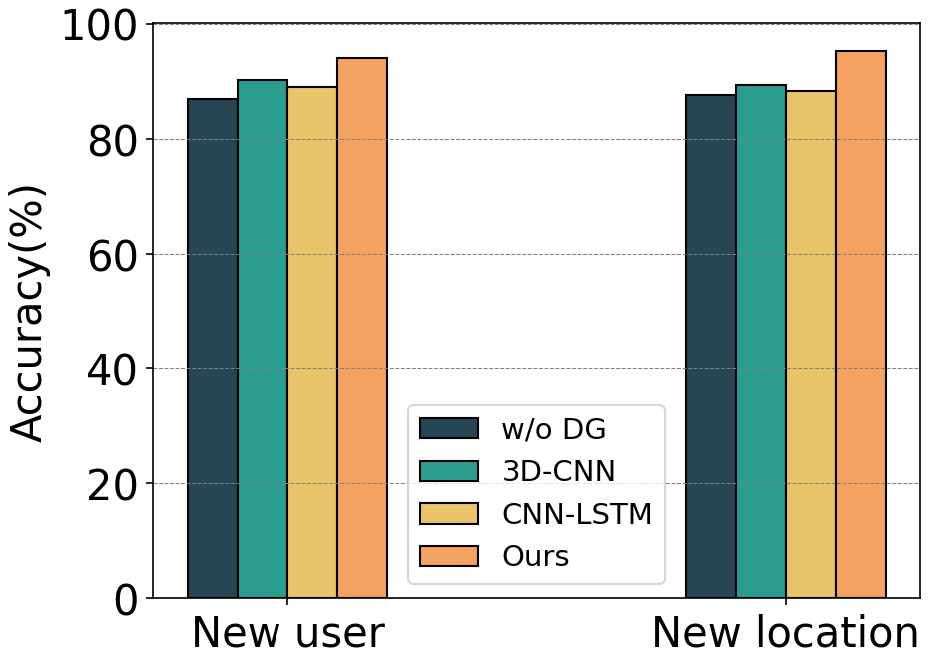}}
\hspace{0.5cm}
\subfigure[Existing works]{
\label{FigMmwaveComparison:b}
\includegraphics[height=2.8cm]{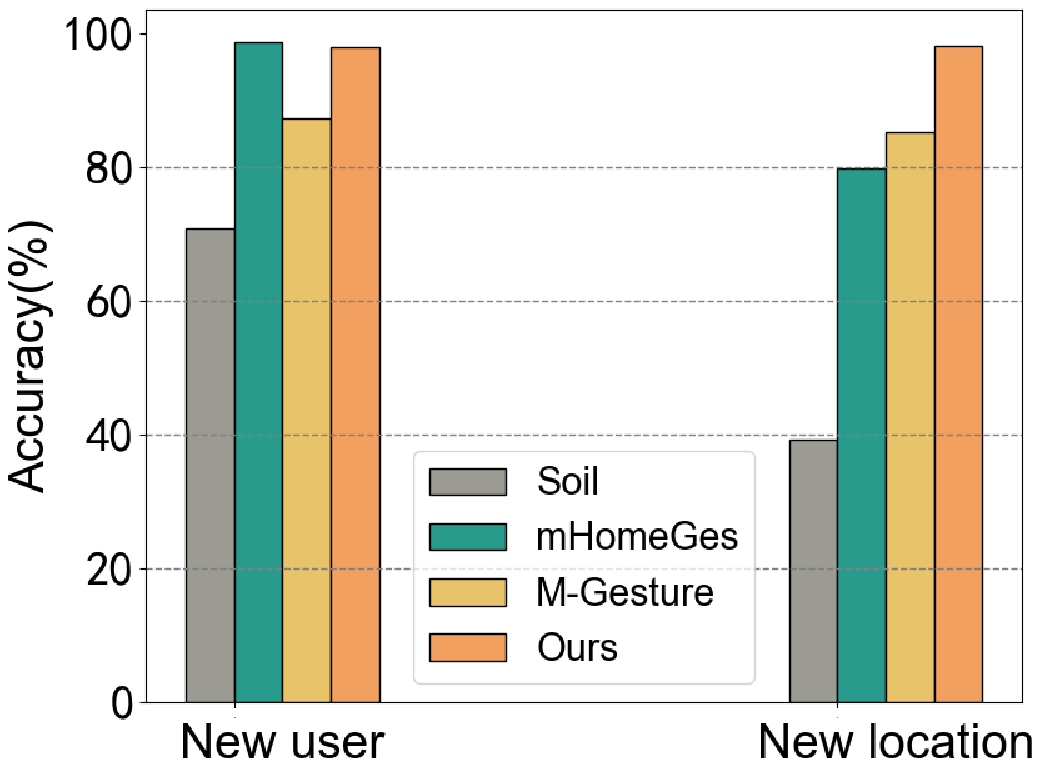}}
\hspace{0.5cm}
\subfigure[Input extraction methods]{
\label{FigMmwaveComparison:c}
\includegraphics[height=2.8cm]{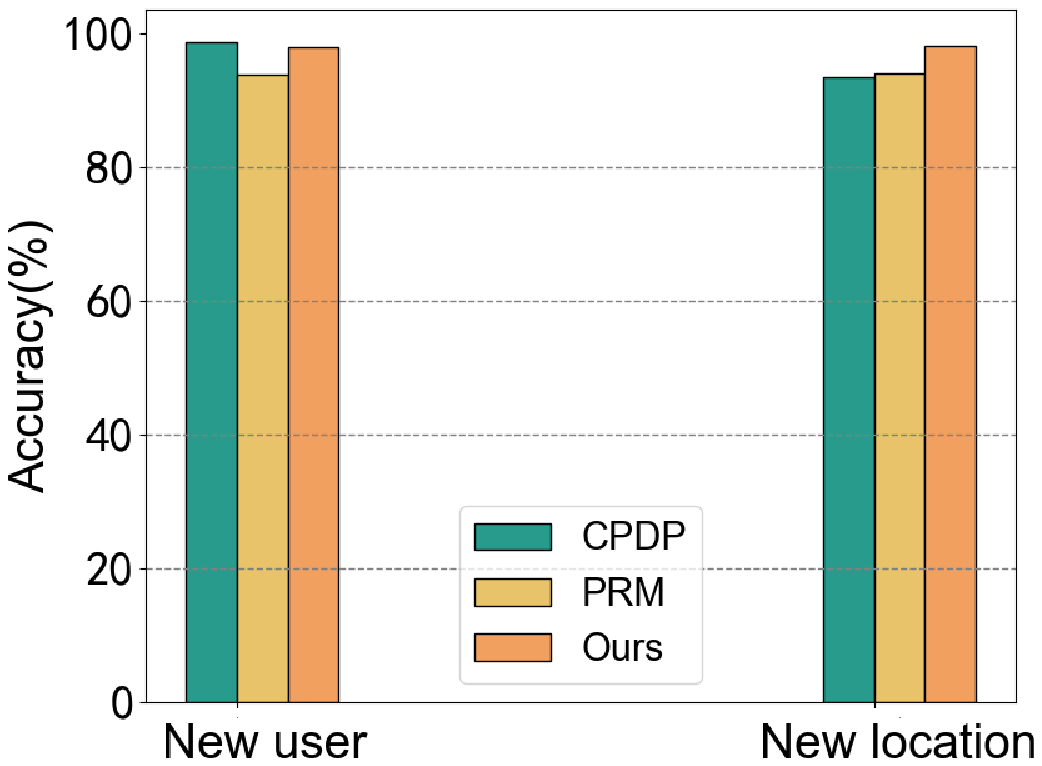}}
\caption{Method comparison in new domains.}
\label{FigMmwaveComparison}
\end{center}
\end{figure*}

\subsubsection{In-domain accuracy}
We first evaluated the in-domain accuracy of the mmWave activity recognition system, involving 6 activities by 6 volunteers at 6 locations. Each activity was repeated 20 times by each volunteer. We conducted 5-fold cross-validation on the dataset and achieved the average accuracy of 99.5\% for familiar users at seen locations.

\subsubsection{Quality of virtual data}
We augmented the training set by single-modal virtual data generation. The number of virtual samples equaled the number of real samples. We conducted experiments to verify the quality of the virtual data. We trained the feature extractor and classifier with the real data and tested on the virtual data, the accuracy achieved 99.7\%, close to the in-domain accuracy of 99.5\%, verifying that the virtual data followed the same distribution with the real data. We then trained the feature extractor and classifier with the virtual data and tested on the real data, the accuracy achieved 94.2\%, showing that the virtual data brought adequate diversity.

\subsubsection{New user}
To evaluate the generalization of the system on new users, we asked the 6 volunteers to perform activities at location 1. Each volunteer was chosen as the new user in turn, while the other 5 volunteers served as the source domains. The activity recognition results of new users are shown in Fig.~\ref{MmwaveActivityResults:a}, where each column represents the activity recognition accuracy with the current user as the new user. Our method demonstrated an average recognition accuracy of 95.4\% for new users, significantly surpassing the average accuracy of 87.1\% attained without virtual data generation and domain generalization.

\subsubsection{New location}
To evaluate the generalization of the method at new locations, we collected data at 6 locations in the laboratory, as marked on Fig.~\ref{MmWaveExperimentSetup:a}. Each location was chosen as the new domain in turn, and the other 5 locations were taken as the source domains. The activity recognition results at new locations are shown in Fig.~\ref{MmwaveActivityResults:b}, where each column represents the recognition accuracy with the current location as the new domain. The average recognition accuracy of our method achieved 95.4\% at new locations, significantly outperforming the average accuracy of 87.8\% obtained without virtual data generation and domain generalization.

\subsubsection{Comparison of learning models}
We compared our method with the \textit{w/o DG} method, the \textit{3DCNN} model by Chen et al.~\cite{ChenH:2022}, and the \textit{CNN-LSTM} model by Zhang et al.~\cite{ZhangG:2020}. The \textit{w/o DG} method excludes virtual data generation and episodic training. The \textit{3DCNN} model~\cite{ChenH:2022} extracts the features from the time-series of range-Doppler maps. The \textit{CNN-LSTM} model~\cite{ZhangG:2020} employs a 2DCNN model to extract the features from each frame, and employs an LSTM model to further capture the relationship between frames and extract the gesture features. The models were compared concerning their generalization to new users and new locations. The results are illustrated in Fig.~\ref{FigMmwaveComparison:a}, in which each column represents the average recognition accuracy achieved by a specific model. Our method achieved higher recognition accuracy than the other models for new users and at new locations. This can be attributed to the fact that our method is a domain generalization approach that integrates virtual data generation and domain independent feature extraction. In contrast, the \textit{3DCNN} model and the \textit{CNN-LSTM} model solely employ source domain data for model training without extracting domain independent features. Although they exhibit high performance in familiar domains, their effectiveness significantly diminishes in unseen domains. 

\subsubsection{Comparison with existing works}
We compared our system with the existing works \textit{Soli}~\cite{Soli:2016}, \textit{mHomeGes}~\cite{LiuH:2020} and \textit{M-Gesture}~\cite{LiuH:2022}. \textit{Soli}~\cite{Soli:2016} was based on Google’s Soli sensor, leveraging a combination of deep convolutional and recurrent neural networks to recognize gestures. \textit{mHomeGes}~\cite{LiuH:2020} recognized continuous arm gestures based on a lightweight CNN using CPDP to represent the unique features from different arm joints. \textit{mHomeGes} could work across various smart-home scenarios regardless of the impact of surrounding interference. \textit{M-Gesture}~\cite{LiuH:2022} incorporated a PRM and a custom-built neural network to depict and extract the inherent gesture features. Their comparison results are illustrated in Fig.~\ref{FigMmwaveComparison:b}, using the datasets across users and locations. \textit{Soli} did not consider the cross domain issue, hence it did not perform well in new domains. Although without specific domain generalization schemes, \textit{mHomeGes} designed a robust feature extraction method and achieved high accuracy for new users, but did not perform well at new locations. \textit{M-Gesture} required about 20 persons to achieve more than 90\% accuracy with 5 gestures for new users in their experiments, hence it did not perform well with only 5 persons in our experiments. And it did not consider the cross location issue. Overall our method performed the best in new domains.

\subsubsection{Comparison of input extraction methods}
For mmWave activity recognition, we use compressed Doppler maps as input to the framework. To prove its effectiveness, we re-implemented the input extraction methods of \textit{CPDP} from \textit{mHomeGes}~\cite{LiuH:2020} and \textit{PRM} from \textit{M-Gesture}~\cite{LiuH:2022}, and integrated them into our DGSense framework. We compared our input extraction method with \textit{CPDP} from \textit{mHomeGes} and \textit{PRM} from \textit{M-Gesture} by experiments, using the datasets across users and locations. The comparison results are shown in Fig.~\ref{FigMmwaveComparison:c}. \textit{CPDP} achieved slightly higher accuracy than ours for new users, but lower accuracy than ours at new locations, which demonstrated that \textit{CPDP} was robust to different users but did not consider the location dependence issue. \textit{PRM} achieved lower accuracy than ours for both new users and new locations, due to the reason that the number of training domains were less than required. The experiments showed that \textit{CPDP} with our framework outperformed \textit{mHomeGes} and \textit{PRM} with our framework outperformed \textit{M-Gesture}, indicating the effectiveness of our domain generalization strategy.

\subsubsection{Computation costs}
The system ran on a desktop equipped with a GPU of Colorful RTX 3090 and a CPU of Intel i5-12600K. For the training of 720 samples (involving 5 users, 6 activities), the system spent 38s for data preprocessing, 387s to train the virtual data generator, 0.3s to generate 720 virtual samples, and 945s for episodic training with real and virtual samples in 5 source domains. In total, it took 1370s (0.38h) for the training process. To recognize an activity, the system spent 63.9ms for data preprocessing, 0.15ms to extract the features and classify the activity. In total, it took 64.1ms to recognize an activity. As we used 2D compressed Doppler maps as input, the training took less than one hour, and the recognition was in real-time.

\section{Evaluations on acoustic fall detection}
\label{SecAcoustic}

\subsection{Acoustic signals and preprocessing}
\begin{figure}
\begin{center}
\subfigure[Laboratory]{
\label{AcousticExperimentSetup}
\raisebox{0.1cm}{\includegraphics[height=2.2cm,width=2cm]{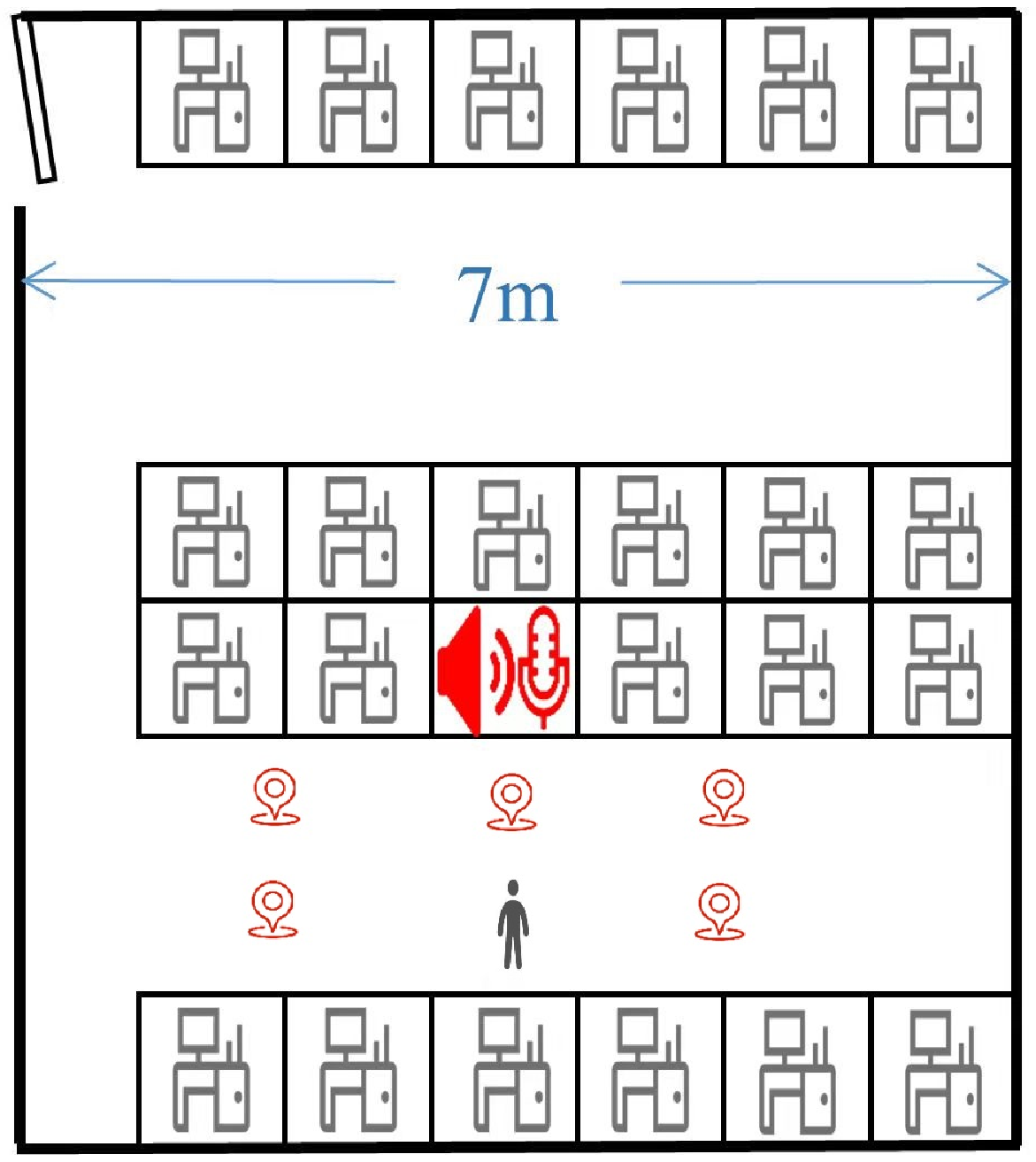}}}
\hspace{0.1cm}
\subfigure[Doppler effect]{
\label{FigDopplerEffect}
\raisebox{0.2cm}{\includegraphics[width=2.3cm]{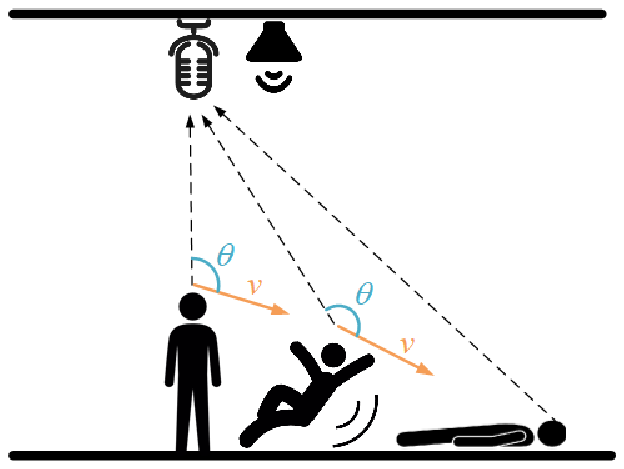}}}
\hspace{0.1cm}
\subfigure[Fall]{
\label{FigAcousticSpectrogramFall}
\includegraphics[width=3.2cm]{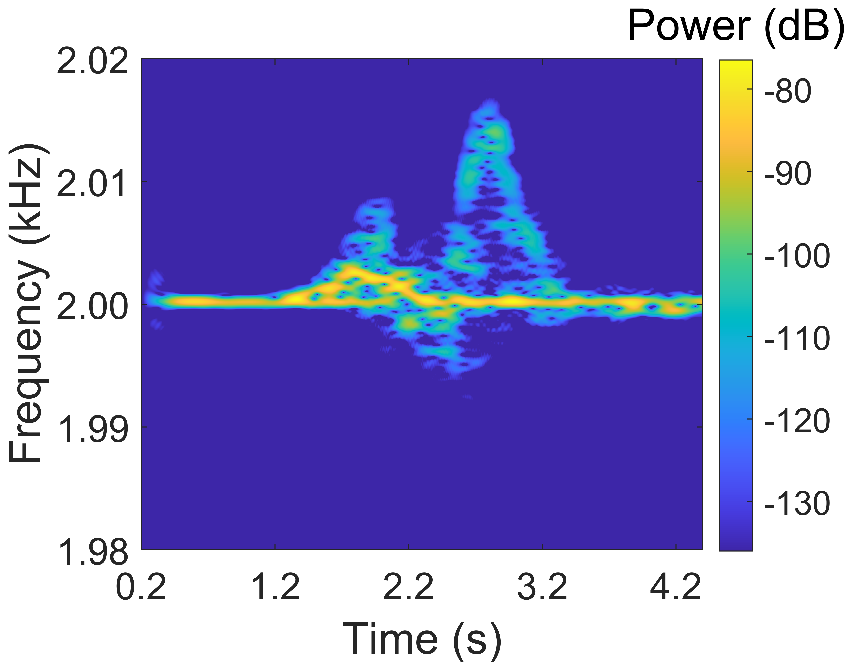}}
\end{center}
\caption{Doppler effect and acoustic spectrograms.}
\label{FigAcousticSpectrogram}
\end{figure}

The fall detection system deploys a speaker and a microphone, binding together as one audio device. The acoustic signals with a fixed frequency are emitted by the speaker, reflected by the human body, and received by the microphone. As human motions can cause Doppler effect on acoustic signals, the frequency of the received acoustic signals may change with the motions, expressed as~\cite{LianJ:2021}:
\begin{equation}
f_r = f_t \cdot \frac{c + v\cdot\cos\theta}{c - v\cdot\cos\theta}
\end{equation}
where $f_t$ is the transmitting frequency, $f_r$ is the receiving frequency, $c$ is the propagation speed of acoustic waves in the air, $v$ is the speed of human body, $\theta$ is the angle between the motion and the beam, as illustrated in Fig.~\ref{FigDopplerEffect}. Different motions usually produce different moving directions and speeds, hence incur different frequency shifts. Applying STFT on the received signals, the Doppler frequency spectrogram can be generated, as shown in Fig.~\ref{FigAcousticSpectrogramFall}. By analyzing the patterns enclosed in the Doppler spectrograms, the motions including falls can be identified. 

\subsection{System design}
The system is based on the DGSense framework, consisting of data collection and preprocessing, virtual data generation, domain independent feature extraction and classification. As the samples are spectrograms, we employ a single-modal virtual data generator. The virtual samples are generated by adding noise to the intermediate features and decoding them. Episodic training is exploited to extract domain independent motion features. The main network and the domain networks share the same structure, comprising a feature extractor and a binary classifier. The feature extractor is based on ResNet18 and the binary classifier is a 3-layer fully connected neural network to classify the motions as falls or non-falls.

\subsection{Evaluations}
The fall detection experiments were conducted in a laboratory, as shown in Fig.~\ref{AcousticExperimentSetup}, with the size of 8m$\times$7m. The equipments consisted of an Edifier R1080BT loudspeaker and a SAMSON Meteomic microphone (16bit, 48kHz). The loudspeaker emitted signals at the frequency of 20kHz, inaudible to human ears. The microphone sampled the signals at the frequency of 48kHz to capture the Doppler effect. Early studies leveraged the sounds caused by falls for detection~\cite{Khan:2015,Principi:2016}, which could be affected by the environmental sounds. Our method detects falls with inaudible acoustic signals, avoiding the sound interference from the surroundings.
We chose three metrics to evaluate the performance of fall detection: accuracy, precision and recall. \textit{Accuracy} is defined as the ratio of correctly identified events out of all events, \textit{precision} is defined as the proportion of actual falls out of detected falls, and \textit{recall} is defined as the proportion of correctly identified falls out of actual falls. High accuracy, precision and recall are preferred.  

\subsubsection{In-domain performance}
We first evaluated the in-domain performance of the acoustic fall detection system, involving 4 volunteers and 6 locations. For each volunteer, each non-fall motion was repeated 20 times and the fall motion was repeated 40 times. At each location, each non-fall motion was repeated 10 times and the fall motion was repeated 20 times. We conducted 5-fold cross-validation on the dataset and achieved the average accuracy, precision and recall of 99.8\%, 100\% and 99.3\% for familiar users at seen locations.

\subsubsection{New user}
To evaluate the generalization of the system on new users, we collected data with 4 volunteers (referred to as P1--P4). They performed motions of squatting down, standing up, walking, standing still, and falling. Each non-fall motion was repeated 20 times per person and the fall motion was repeated 40 times per person. We selected 3 volunteers as the source domains and tested on the remaining 1 volunteer as the new domain. As the number of falls and non-falls was imbalanced, we generated 40 virtual fall samples for each volunteer. As a result, the training set had a total of 480 samples, including 120 real fall samples, 120 virtual fall samples, and 240 real non-fall samples. 
The experimental results are shown in Table~\ref{TblAcousticUser}. For new users, our method exhibited an average accuracy of 95.0\%, an average precision of 95.5\%, and an average recall of 94.8\%. 

\subsubsection{New location}
To evaluate the generalization to new locations, we collected data at 6 locations (referred to as L1--L6), as marked on Fig.~\ref{AcousticExperimentSetup}, with adjacent locations 1m apart. The motions were squatting down, standing up, walking, jumping, and falling. Each non-fall motion was repeated 10 times at each location and the fall motion was repeated 20 times at each location. We took 4 locations as the source domains and tested at the remaining 2 locations as the new domains. As the number of falls and non-falls was imbalanced, we generated 20 virtual fall samples at each location. Consequently, the training set had a total of 320 samples, including 80 real fall samples, 80 virtual fall samples, and 160 real non-fall samples. 
The results are shown in Table~\ref{TblAcousticLocation}. At new locations, our method achieved the average accuracy of 95.3\%, the average precision of 96.3\%, and the average recall of 94.3\%. 

\subsubsection{Comparison with existing works}
To prove the effectiveness of our acoustic fall detection system, we compared it with the work proposed by Lian et al.~\cite{LianJ:2021} and without domain generalization (abbr. w/o DG). Lian et al.~\cite{LianJ:2021} extracted a set of effective features (the speed feature, the extreme ratio curve, and the spectral entropy) from the acoustic Doppler frequency spectrograms, applied SVD to reduce the dimension, and leveraged HMM to classify motions as falls or non-falls. We re-implemented the feature extraction method of \cite{LianJ:2021} and leveraged an SVM to perform binary classification. The comparisons are listed in Table~\ref{TblAcousticComp}. Our method notably outperformed w/o DG for new users and at new locations. The method of \cite{LianJ:2021}+SVM could reduce the influence of environmental noise and showed high accuracy for new users. But they did not consider the issue of location dependence, hence did not perform well at new locations. Overall, for in-domain performance, our method outperformed \cite{LianJ:2021}+SVM. For cross domain performance, our method was comparable with \cite{LianJ:2021}+SVM for new users and outperformed \cite{LianJ:2021}+SVM at new locations. 

\subsubsection{Computation costs}
The system ran on a desktop equipped with a GPU of Colorful RTX 3090 and a CPU of Intel i5-12600K. For the training of 360 samples involving 3 volunteers, the system spent 257s for data preprocessing, 344s to train the virtual data generator, 0.2s to generate 120 virtual samples, and 3035s for episodic training in 3 source domains. In total, it took 3636s (1.01h) for the training process. To detect a fall or non-fall motion, the system spent 847ms to preprocess the sample, 2.6ms to extract the features and classify the motion. In total, it took 849.6ms to make a fall detection. As the acoustic signals were sampled at the frequency of 48KHz in order to capture the Doppler effect, the preprocessing took a longer time to generate the Doppler frequency spectrogram by STFT. But 849.6ms was still a real-time response.

\begin{table}
		\footnotesize
    \caption{Acoustic fall detection on new users (unit: \%).}
    \label{TblAcousticUser}
    \centering
    \begin{tabular}{ccccc}
    \hline
        Source domain & Target domain & Accuracy & Precision & Recall \\ \hline
        P2, P3, P4 		& P1 						& 95 	& 93 	& 98 \\ 
        P1, P3, P4 		& P2 						& 94 	& 95 	& 93 \\ 
        P1, P2, P4 		& P3 						& 95 	& 97 	& 93 \\ 
        P1, P2, P3 		& P4 						& 96 	& 97 	& 95 \\ \hline
        \multicolumn{2}{c}{Average} 	& \textbf{95.0} 	& \textbf{95.5} 	& \textbf{94.8} \\ \hline
    \end{tabular}
\end{table}

\begin{table}
    \centering
		\footnotesize
    \caption{Acoustic fall detection at new locations (unit: \%).}
    \begin{tabular}{ccccc}
    \hline
        Source domain 	& Target domain & Accuracy & Precision 	& Recall \\ \hline
        L3, L4, L5, L6 	& L1, L2				& 95 		& 95 		& 95 \\
        L1, L2, L5, L6 	& L3, L4				& 96 		& 97 		& 95 \\ 
        L1, L2, L3, L4 	& L5, L6				& 95 		& 97 		& 93 \\  \hline
        \multicolumn{2}{c}{Average}			& \textbf{95.3} 	& \textbf{96.3} 	& \textbf{94.3} \\	\hline
    \end{tabular}
    \label{TblAcousticLocation}
\end{table}

\begin{table*}
    \centering
		\footnotesize
    \caption{Comparison with existing works (unit: \%).}
    \begin{tabular}{c|ccc|ccc|ccc}
    \hline
										& \multicolumn{3}{|c|}{In-domain} & \multicolumn{3}{|c|}{New user} 	& \multicolumn{3}{|c}{New location}  \\ 
										& Accuracy & Precision & Recall 	& Accuracy & Precision & Recall & Accuracy & Precision & Recall 		\\ \hline
		w/o DG  											& 99.8 & 100 	& 99.3 		& 87.4 & 87.5 & 87.5			& 89.1 & 88.5 & 90.0  \\
    \cite{LianJ:2021}+SVM					& 95.7 & 97.4 & 89.6		& 96.5 & 99.4 & 90.0 			& 90.8 & 79.7 & 97.5 \\ 
    Ours													& 99.8 & 100 	& 99.3		& 95.0 & 95.5 & 94.8			& 95.3 & 96.3	& 94.3  \\  \hline
    \end{tabular}
    \label{TblAcousticComp}
\end{table*}

\section{Ablation study}
\label{SecAblationStudy}

\subsection{Cross-modal vs. multi-modal generator}
To validate the effect of the cross-modal virtual data generator, we conducted experiments to compare it with the multi-modal virtual data generator. We used the WiFi gesture datasets in the laboratory (R2) and the small meeting room (R4) and compared their accuracy for new users. The experimental results are presented in Table~\ref{TblVirtualDataGeneration}. In WiFi sensing, the amplitude, the phase and the spectrogram have inter-modal relationships. Due to the inconsistency introduced by adding noise to each modality respectively, the multi-modal virtual data generator did not performance well. The cross-modal virtual data generator significantly improved the ability of generalization to new users. 

\begin{table}[h]
		\footnotesize
    \centering
		\caption{Cross-modal vs. multi-modal generator.}
		\label{TblVirtualDataGeneration}
    \begin{tabular}{p{2cm}cc}
    \hline
		Generator				& New user in R2 		& New user in R4 	\\  \hline
    w/o generator 	& 74.8\% 						& 78.4\% 					\\ 
		Multi-modal  		& 73.8\%						& 76.5\%					\\
    Cross-modal  		& 84.6\% 						& 86.9\% 					\\ \hline  
		\end{tabular}
\end{table}

\begin{table}[h]
		\footnotesize
    \centering
		\caption{Methods of virtual data generation.}
		\label{TblGenerationMethod}
    \begin{tabular}{p{3cm}cc}
    \hline
		Generator							& New user	& New location		\\ \hline
    w/o generator 				& 91.4\% 			& 92.1\% 			\\ 
    Autoencoder-based 		& 95.3\% 			& 96.7\% 			\\ 
    GAN-based 						& 96.5\%			& 96.1\% 			\\ 
    VAE-based (ours)			& 97.5\% 			& 97.2\% 			\\ \hline
    \end{tabular}
\end{table}

\begin{table}[h]
		\footnotesize
    \centering
		\caption{Methods of domain generalization.}
		\label{TblGeneralizationMethod}
    \begin{tabular}{p{3cm}cc}
    \hline
    Method 							& New user		& New location		\\ \hline
    w/o generalization 	& 91.7\% 			& 90.8\% 			\\ 
    GAN (25\%) 					& 89.6\% 			& 89.5\% 			\\ 
    GAN (50\%) 					& 95.3\% 			& 96.4\% 			\\ 
    MMD (25\%) 					& 93.1\% 			& 94.9\% 			\\ 
    MMD (50\%) 					& 94.3\% 			& 95.2\% 			\\ 
    DGSense (ours) 			& 97.5\% 			& 97.2\% 			\\ 		\hline
    \end{tabular}
\end{table}

\subsection{Method of virtual data generation}
To design an effective virtual data generator, the choice of the generating method is important. We chose VAE as the generating method. To demonstrate its effect, we compared it with the Autoencoder-based generator and the GAN-based generator, by experiments on mmWave activity recognition, using the datasets across users and locations. The experimental results are shown in Table~\ref{TblGenerationMethod}. All the virtual data generators improved the accuracy, as they all enlarged the datasets and brought diversity, while the VAE-based generator achieved the best performance by a small margin.

\subsection{Method of domain generalization}
DGSense exploits episodic training to achieve domain generalization. We compared it with GAN-based and MMD-based domain adaptation schemes, by experiments on mmWave activity recognition across users and locations. The GAN-based scheme regards the target domain data as fake samples and extracts the domain independent features through adversarial training. The MMD-based scheme shrinks the distribution discrepancy between the domains in Reproducing Kernel Hilbert Space (RKHS). GAN-based and MMD-based schemes are domain adaptation methods, so they require some unlabeled data from the target domains. In the experiments, they took 25\% and 50\% unlabeled target domain data for training, while our method did not use any target domain data. The results are shown in Table~\ref{TblGeneralizationMethod}. Our method achieved the best performance. The GAN-based scheme required certain amount of unlabeled target domain data to achieve improvement. It exhibited great improvement with 50\% unlabeled target domain data, but no improvement with only 25\%. The MMD-based scheme achieved domain adaptation by feature alignment, hence its improvement was not much. 

\subsection{Number of source domains}
To investigate the impact of the number of source domains on the performance, we conducted WiFi gesture recognition experiments in the laboratory (R2). We took 5, 4, 3, 2, 1 users as the source domains and the remaining 1, 2, 3, 4, 5 users as the unseen target domains. The number of real and virtual samples per gesture per source domain was 20 respectively. Fig.~\ref{FigNumDomain} shows the results. The accuracy increased with the number of source domains. With 4 source domains, the method could achieve the accuracy of 82.5\%. With 5 source domains, the method could achieve the accuracy of 84.6\%. The cost of more source domains was more training time. From 1 to 5 source domains, the episodic training time increased from about 0.3h to about 3.5h. Hence, the number of source domains is important for the generalization performance and 4--5 source domains can achieve proper performance.

\subsection{Number of real samples}
To investigate the impact of the number of real samples on the performance, we conducted WiFi gesture recognition experiments in the laboratory (R2). We took 5 users as the source domains and the remaining 1 user as the unseen target domain. The number of real samples per gesture per source domain ranged from 4,8,12,16 to 20. We generated the same number of virtual samples. Fig.~\ref{FigNumSample} shows the results. The accuracy increased with the number of real samples. With 16 real samples per gesture per domain, the method could achieve the accuracy of 80.5\%. With 20 real samples, the method could achieve the accuracy of 84.6\%. The cost of more samples was more training time. From 4 to 20 real samples per gesture per domain, the episodic training time increased from about 0.7h to about 3.5h. Hence, the number of real samples is important for the generalization performance and 16--20 real samples per gesture per domain can achieve proper performance.

\subsection{Number of virtual samples}
\begin{figure*}
\begin{center}
\subfigure[Source domains]{
\label{FigNumDomain}
\includegraphics[width=3.5cm]{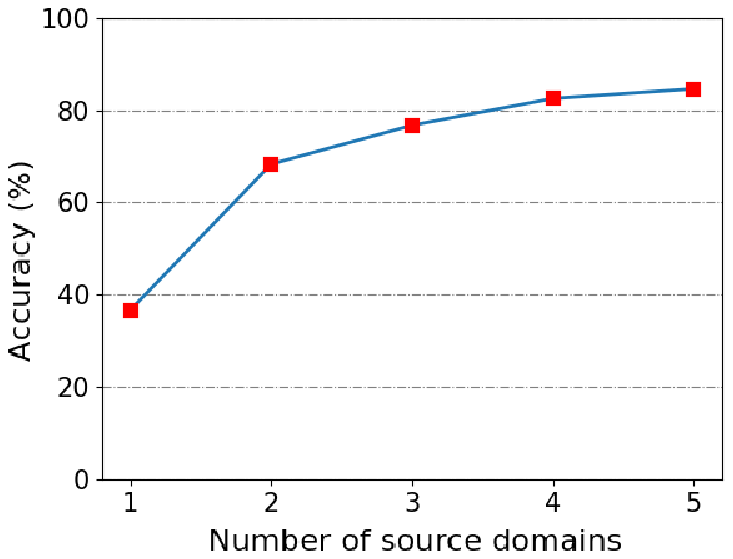}}
\hspace{0.5cm}
\subfigure[Real samples]{
\label{FigNumSample}
\includegraphics[width=3.5cm]{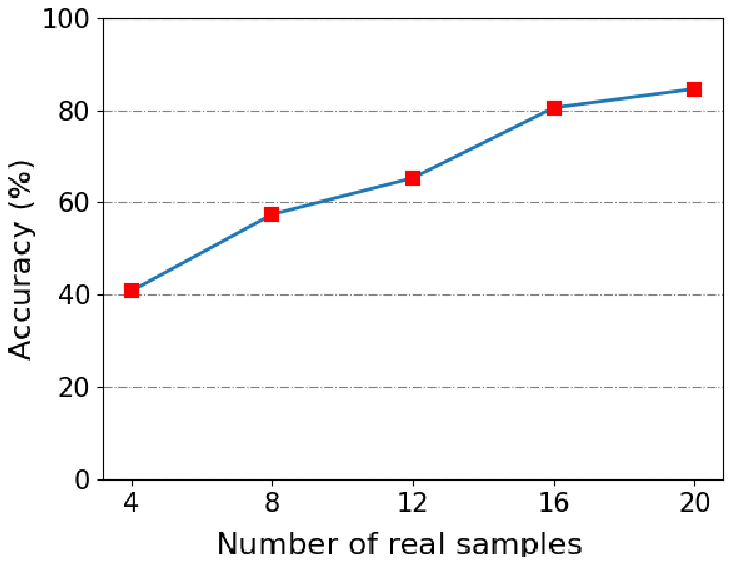}}
\hspace{0.5cm}
\subfigure[Virtual samples]{
\label{FigNumVirtual}
\includegraphics[width=3.5cm]{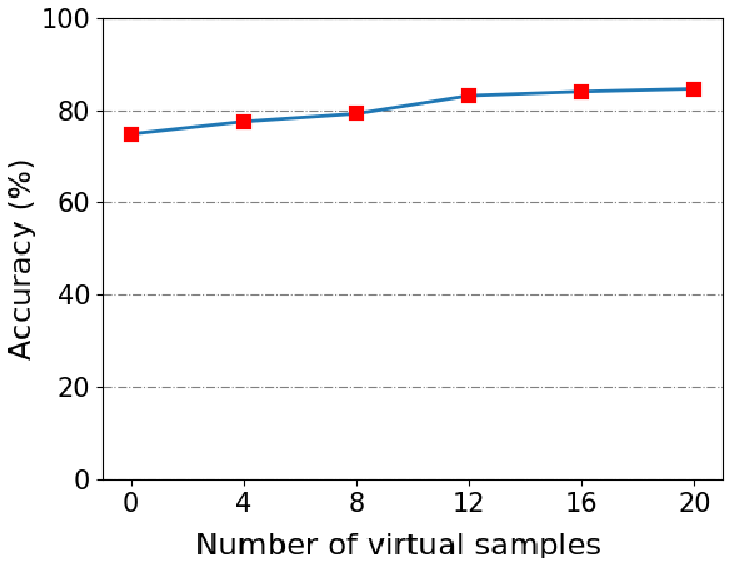}}
\end{center}
\caption{Impact of number of source domains and samples.}
\label{FigParameter}
\end{figure*}

We also investigated the impact of the number of virtual samples on the performance. To this end, we conducted WiFi gesture recognition experiments in the laboratory (R2). We took 5 users as the source domains and the remaining 1 user as the unseen target domain. Per gesture per source domain, the number of real samples was 20, the number of virtual samples ranged from 4,8,12,16 to 20. Fig.~\ref{FigNumVirtual} shows the results. The accuracy increased slightly with the number of virtual samples. When the virtual samples increased from 4 to 20, the accuracy increased from 77.5\% to 84.6\%, and the episodic training time increased from about 2h to about 3h. Hence, incorporating virtual samples helps to improve the performance.

\section{Discussions}
\label{SecDiscussions}

The proposed framework DGSense can effectively boost the generalization of wireless sensing to unseen domains. All of our experiments were conducted in real-world scenarios, hence our framework can be applied in practice. The sensing devices are commodity devices, which are easy to obtain. Notably, there are a few points that we should pay attention to during the application of DGSense in practice. Firstly, different sensing mediums have different sensing signals, resulting in different data dimensions and characteristics. Under the framework of DGSense, the preprocessing components may leverage different data processing techniques and the feature extractors may have different learning structures. Secondly, although our method was evaluated with only 6 gestures or activities, it can be extended to recognize more gestures and activities. Thirdly, to train the domain generalization model, multiple source domains are required. More source domains will help improve the generalization ability, but incur more training time. In our experiments, 4--5 source domains were adequate. Finally, although we evaluated only on WiFi gesture recognition, mmWave activity recognition and acoustic fall detection, the framework is applicable to other wireless sensing tasks. 

There are a few limitations in the current form of our work, which will be the directions of our further investigation. Firstly, our method is designed to perform recognition within the predefined gestures or activities. For real-world applications, there may be unseen gestures or activities, which may lead to erroneous predictions. Currently we regard the unseen gestures or activities as outliers and detect them by an outlier detector. Secondly, our method is for single-person scenarios. It can not handle multi-person scenarios yet. When multiple persons perform gestures or activities simultaneously, our method still treats them as one person, leading to erroneous predictions. We leave the multi-person sensing issue as our next step work. 

\section{Conclusions}
\label{SecConclusions}
To achieve domain independent wireless sensing, we propose a general framework of domain generalization---DGSense. The framework comprises four components: data collection, data preprocessing, virtual data generation, domain independent feature extraction and classification. The key points lie in virtual data generation and domain independent feature extraction. According to the modalities in the data, we devise either single-modal or cross-modal virtual data generators to augment the training set and increase the data diversity to enhance the robustness of the sensing model. To achieve domain independence, we tailor the feature extractors to the wireless signals and leverage episodic training on the main network and the domain networks to extract domain independent features. 

The framework is a general solution adaptable to diverse wireless sensing tasks on diverse wireless signals. To prove its generality and effectiveness, we developed a WiFi gesture recognition system, an mmWave activity recognition system, and an acoustic fall detection system, all based on DGSense. 
For WiFi gesture recognition, we took CSI amplitude, phase and Doppler spectrogram as the input. We set amplitude as the base modality and generated virtual amplitude, phase and spectrogram through the cross-modal virtual data generator. We employed 1DCNN to extract features from amplitude and employed ResNet18 to extract features from phase and spectrogram. Through episodic training, the main network gained the capability of extracting domain independent gesture features. 
For mmWave activity recognition, we compressed 3D multi-frame range-Doppler maps to 2D Doppler maps as the input, and generated virtual 2D Doppler maps with the single-modal virtual data generator. We employed ResNet18 coupled with CBAM as the feature extractor. Through episodic training, the main network could extract domain independent activity features. 
For acoustic fall detection, we transformed acoustic waves to Doppler spectrogram as the input. Virtual spectrograms were generated by the single-modal virtual data generator. We employed ResNet18 as the feature extractor. Through episodic training, the main network attained the capability of extracting domain independent fall features. 

Real-world evaluations on the three systems showcased the generality of DGSense across different sensing tasks on different wireless signals. The experimental results demonstrated the generalization capability of DGsense to unseen domains, including new users, new locations and new environments.

\bibliographystyle{IEEEtran}
\bibliography{references}

\end{document}